\documentclass{article}

    \PassOptionsToPackage{numbers, compress}{natbib}
\usepackage[preprint]{neurips_2026}

\usepackage[utf8]{inputenc} 
\usepackage[T1]{fontenc}    
\usepackage{hyperref}       
\usepackage{url}            
\usepackage{booktabs}       
\usepackage{amsfonts}       
\usepackage{nicefrac}       
\usepackage{microtype}      
\usepackage{xcolor}         
\usepackage{amssymb}
\usepackage{hyperref}
\usepackage{url}
\usepackage{graphicx}
\usepackage{subcaption}
\usepackage{amsthm}
\usepackage{algorithm}
\usepackage{algorithmic}
\usepackage{booktabs}
\usepackage{wrapfig}
\usepackage{amsmath}
\newtheorem{proposition}{Proposition}
\newtheorem{lemma}{Lemma}
\newtheorem{corollary}{Corollary}
\usepackage{multirow}

\title{GR4CIL: Gap-compensated Routing for CLIP-based\\ Class Incremental Learning}

\author{%
  Tianqi Wang \\
  Department of COMP\\
  The Hong Kong Polytechnic University\\
  Hong Kong SAR \\
  \And
  Jingcai Guo\thanks{Corresponding author: Jingcai Guo.} \\
  Department of COMP/LSGI \\
  The Hong Kong Polytechnic University\\
  Hong Kong SAR \\
}

\begin{document}

\maketitle

\begin{abstract}

  Class-Incremental Learning (CIL) aims to continuously acquire new categories while preserving previously learned knowledge. Recently, Contrastive Language-Image Pre-trained (CLIP) models have shown strong potential for CIL due to their powerful generalization ability. However, existing methods still face two key challenges: shared-parameter adaptation tends to cause old-knowledge drift, and task-specific knowledge organization often leads to poorly calibrated cross-task responses, making reliable routing difficult. To address these issues, we propose GR4CIL, a framework combining task discrimination and knowledge routing for CLIP-based CIL. GR4CIL preserves task-specific visual knowledge while maintaining an incrementally stable shared textual semantic space, thereby reducing interference across tasks. Moreover, we introduce an orthogonal compensation mechanism to mitigate modality-gap-induced bias, enhance within-task discrimination, and enlarge the score margin between the ground-truth task and competing tasks. As a result, GR4CIL enables more reliable task-aware routing over learned knowledge while retaining the zero-shot generalization capability. Experiments on multiple benchmarks show that GR4CIL consistently outperforms strong baselines.
  

\end{abstract}

\section{Introduction}
\label{intorduction}

Class-Incremental Learning aims to enable models to continuously evolve their knowledge, while preserving previously acquired capabilities~\citep{li2024CLDNet, li2024harnessing}. In recent years, pre-trained vision-language models, particularly CLIP~\citep{radford2021learning}, have increasingly been regarded as a promising foundation for CIL due to their strong cross-modal semantic priors and generalization ability~\citep{huang2025mind, yu2024boosting, zhou2025external, luo2025lada, huang2024class}. However, CLIP’s powerful pre-trained capability does not imply that it can be seamlessly applied to CIL~\citep{jha2024clap4clip}. To adapt to downstream incremental tasks, the model must continuously absorb new knowledge while mitigating catastrophic forgetting~\citep{french1999catastrophic}, while retaining its zero-shot transfer ability as much as possible~\citep{liu2025c}. To fully realize CLIP in CIL scenarios, two closely related challenges remain insufficiently addressed.

\begin{figure}[t]
  \centering
  \begin{subfigure}[t]{0.9\linewidth}
    \includegraphics[width=\linewidth]{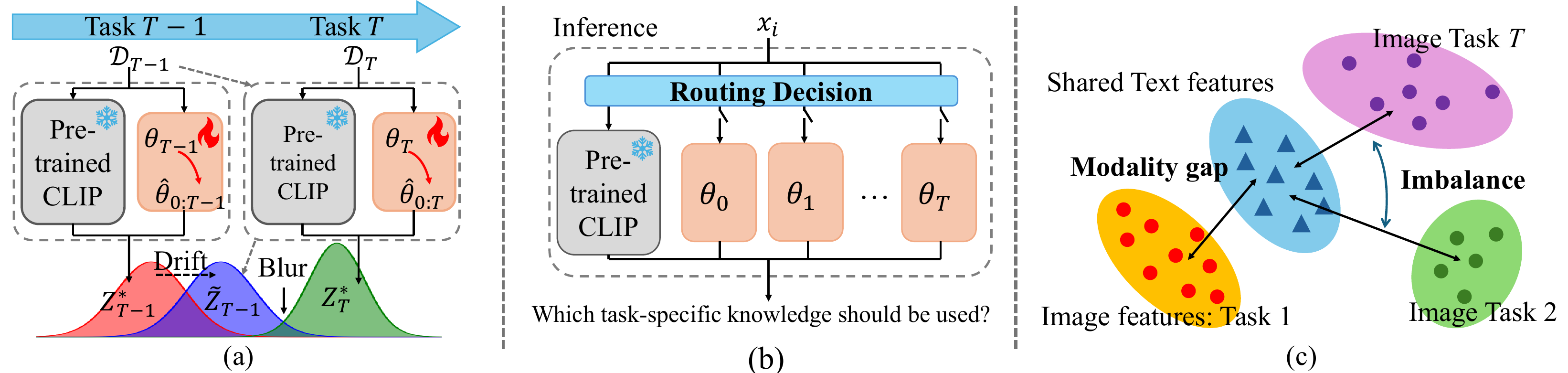}
  \end{subfigure}
  \vskip -0.1in
  \caption{(a) Continual updates of shared parameters cause interference and blur knowledge boundaries. (b) Organizing knowledge into task-specific modules turns inference into a routing problem. (c) The modality gap in CLIP affects both intra-task discrimination and inter-task routing.}
  \vskip -0.22in
  \label{fig1}
\end{figure}

The first challenge arises from the continual update of shared parameters. Existing CIL methods typically rely on either full or parameter-efficient fine-tuning to adapt to new tasks~\citep{huang2025mind, zhou2025external, luo2025lada, gao2024clip, zhou2022learning, wu2025sdlora}. When learnable parameters are repeatedly used across incremental tasks, the optimization of new tasks inevitably alters the representation structures on which old tasks depend~\citep{ijcai2025p715}, thereby blurring the knowledge boundaries (see Fig.~\ref{fig1}(a)). Therefore, a natural idea is to proactively avoid mutual interference among tasks by separately accommodating task-specific knowledge in different modules.


The second challenge lies in inference once knowledge is organized into task-specific structures. After learning task-specific knowledge, the model must determine which learned knowledge should be invoked for a given input without task identity~\citep{yu2024boosting, zhang2024continual, zhang2025visual}. This turns inference into a routing problem and requires clearer discrimination among competing task-specific modules (see Fig.~\ref{fig1}(b)). At the same time, the incremental adaptation should preserve CLIP’s zero-shot capability as much as possible, so that the learned model remains extensible to broader inference scenarios.~\citep{2025arXiv250910535L}.




To address the above challenges, we organize knowledge into task-specific modules during incremental training, and enable knowledge routing during inference with the aid of an auxiliary out-of-distribution (OOD) detection mechanism~\citep{kim2022theoretical}. Specifically, each task branch should respond strongly to its in-distribution samples while suppressing samples from other tasks, which can be interpreted from a task-relative OOD perspective. However, the effectiveness of this paradigm depends not only on whether each task-specific module can provide reliable intra-task discrimination, but also on whether the ground-truth (GT) task can be sufficiently distinguished from competing task-specific knowledge for the same sample during inference~\citep{ijcai2025p715}.


Existing study~\citep{huang2025mind} has shown that the intrinsic modality gap in CLIP can affect performance in CIL. Specifically, text and image features typically lie in two separated narrow cones in the feature space, leading to a modality gap~\citep{liang2022mind}. Under continual adaptation, this gap is further perturbed, which limits the discriminative capability of the text classifier within each task. Moreover, we observe that the modality gaps formed after CIL vary across tasks (see Fig.~\ref{fig1}(c) and Sec.~\ref{Preliminaries}), causing different task-specific modules to respond unevenly to the same sample. Therefore, from the perspective of modality gap, the key is to introduce a compensation mechanism that strengthens intra-task discrimination and enlarges the advantage of the GT task over competing tasks, enabling more reliable routing.


We propose Gap-compensated Routing for CLIP-based CIL (GR4CIL). Specifically, GR4CIL equips the visual branch with task-specific modules, so as to preserve task-exclusive knowledge. On the text branch, GR4CIL learns a shared module to maintain an incremental stable semantic space. Furthermore, we introduce an orthogonal compensation mechanism that compensates image representations in the orthogonal complement of the text space, thereby enhancing intra-task discriminability and enlarging the response margin between the GT task and competing tasks. Finally, GR4CIL further incorporates prototype-driven OOD detection to perform task-aware routing. Moreover, GR4CIL leaves a practical interface for extending inference beyond the standard CIL, where learned knowledge may become insufficient and zero-shot generalization can be invoked. Our contributions are threefold:

{
  \setlength{\leftmargini}{10pt}
  \begin{itemize}

    \item We propose a replay-free framework for CLIP-based CIL that jointly addresses task-specific knowledge learning and task-aware routing. The framework structurally reduces mutual interference among different tasks, while leaving a practical interface for more open inference scenarios.
    

    \item We design an orthogonal compensation mechanism based on the modality gap to improve intra-task discriminability and increase the response margin between the GT task and competing tasks. We further provide a geometric interpretation and theoretical support for this design.
    

    \item Extensive experiments on multiple benchmarks demonstrate that our method consistently outperforms existing approaches without requiring replay samples.
    
  \end{itemize}
}

\section{Related Work}
\label{related_work}


\textbf{CLIP-based CIL.} Existing efforts on CIL of pre-trained models generally follow two technical routes: one adapts the model through full fine-tuning~\citep{zhang2023slca}, while the other performs parameter-efficient fine-tuning by introducing lightweight modules~\citep{wu2025sdlora, wang2022dualprompt, smith2023coda, wang2022learning}. In CLIP-based CIL, prior works have largely followed this line of development, while further incorporating distillation, prototype constraints, or feature fusion to mitigate forgetting. Specifically, ZSCL distills the model with additional data~\citep{zheng2023preventing}. PROOF enhances learning performance through a feature projection module and cross-modal fusion~\citep{zhou2025learning}. LADA designs an expandable adapter and combines it with feature distillation~\citep{luo2025lada}. Magmax progressively fine-tunes the full model while using task vectors for fusion~\citep{marczak2024magmax}. Although these methods have made progress in alleviating forgetting, most of them are still built upon continual updates of shared parameters, making it difficult to avoid inter-task interference and knowledge drift.



\textbf{Task Inference in CIL.} When modules are learned for different tasks, the model must determine which task-specific knowledge should be invoked during inference. In such scenarios, task inference is closely related to OOD detection~\citep{kim2022theoretical}, since the model needs to decide whether an input should be handled by one of the learned tasks or regarded as outside the current knowledge scope~\citep{ijcai2025p715, morteza2022provable, ming2023cider, Lu2024PALM, Kim2022CLOM, kim2022MORE, lin2024TPL}. In CLIP-based CIL, this direction remains relatively underexplored. Recently, MOE4CL learns a set of experts and performs routing with autoencoders~\citep{yu2024boosting}, while LGVLM and AdapterVLM exploit OOD anchors and samples to support inference~\citep{zhang2024continual, zhang2025visual}. However, these methods mainly emphasize task-internal module design, while leaving relatively underexplored how to improve the discriminability and score separability of task-specific knowledge, which is crucial for reliable routing.

\textbf{Modality Gap.} Prior work~\citep{liang2022mind} has shown that CLIP exhibits a modality gap, where text and image features tend to form two separated narrow cones in the shared feature space. Recent studies have explored reducing this gap to improve downstream cross-modal performance~\citep{eslami2025mitigate, mistretta2025cross, schrodi2025two, cai2025misalignment, yamaguchi2025post}. In CLIP-based CIL, MG-CLIP investigates continual adaptation from the perspective of modality gap and uses it to regulate fine-tuning strength~\citep{huang2025mind}. In contrast, our focus is not to directly shrink the gap itself, but to examine how modality-gap-induced residual bias under continual adaptation limits intra-task discrimination and weakens the separability between the GT task and competing tasks during routing.

\section{Preliminaries}
\label{Preliminaries}

\subsection{CLIP-based CIL Problem Setup}

We consider a CIL task sequence based on the CLIP model, denoted as $\{\mathcal{T}^{1}, \mathcal{T}^{2}, \dots, \mathcal{T}^{T}\}$. The $t$-th task consists of a training set $\mathcal{D}^{t}=\{(\mathbf{x}_i^{t}, y_i^{t})\}_{i=1}^{N_t}$ and a class set $\mathcal{C}^{t}$, where the classes of different tasks are mutually disjoint, i.e., $\mathcal{C}^{i}\cap\mathcal{C}^{j}=\varnothing$ for all $i\neq j$~\citep{ijcai2025p715}. During training on task $t$, the model only has access to the current task data $\mathcal{D}^{t}$, and task identifiers are unavailable at test time. In this paper, we adopt a pre-trained CLIP model $M=(f_v, f_t)$ as the backbone, where $f_v$ and $f_t$ denote the visual encoder and the text encoder. Given an input image $\mathbf{x}_i$ and a class prompt $\mathbf{p}(c)$, the visual and textual features are represented as $\mathbf{v}_i=f_v(\mathbf{x}_i)$ and $\mathbf{t}_{c}=f_t(\mathbf{p}(c))$. For decision, the cosine similarity score of sample $\mathbf{x}_i$ for class $c$ is defined as $s^c(\mathbf{x}_i)=\langle \mathbf{v}_i, \mathbf{t}_{c}\rangle$. After training on the first $t$ tasks, the model is required to recognize all seen classes $\mathcal{C}^{1:t}=\bigcup_{i=1}^{t}\mathcal{C}^{i}$ while preserving historical knowledge.


\subsection{Task-wise Inconsistency of Modality Gaps}
\label{insight}

\begin{wrapfigure}[20]{r}{0.32\textwidth}
    \centering
    \vskip-1em
    \includegraphics[width=0.29\textwidth]{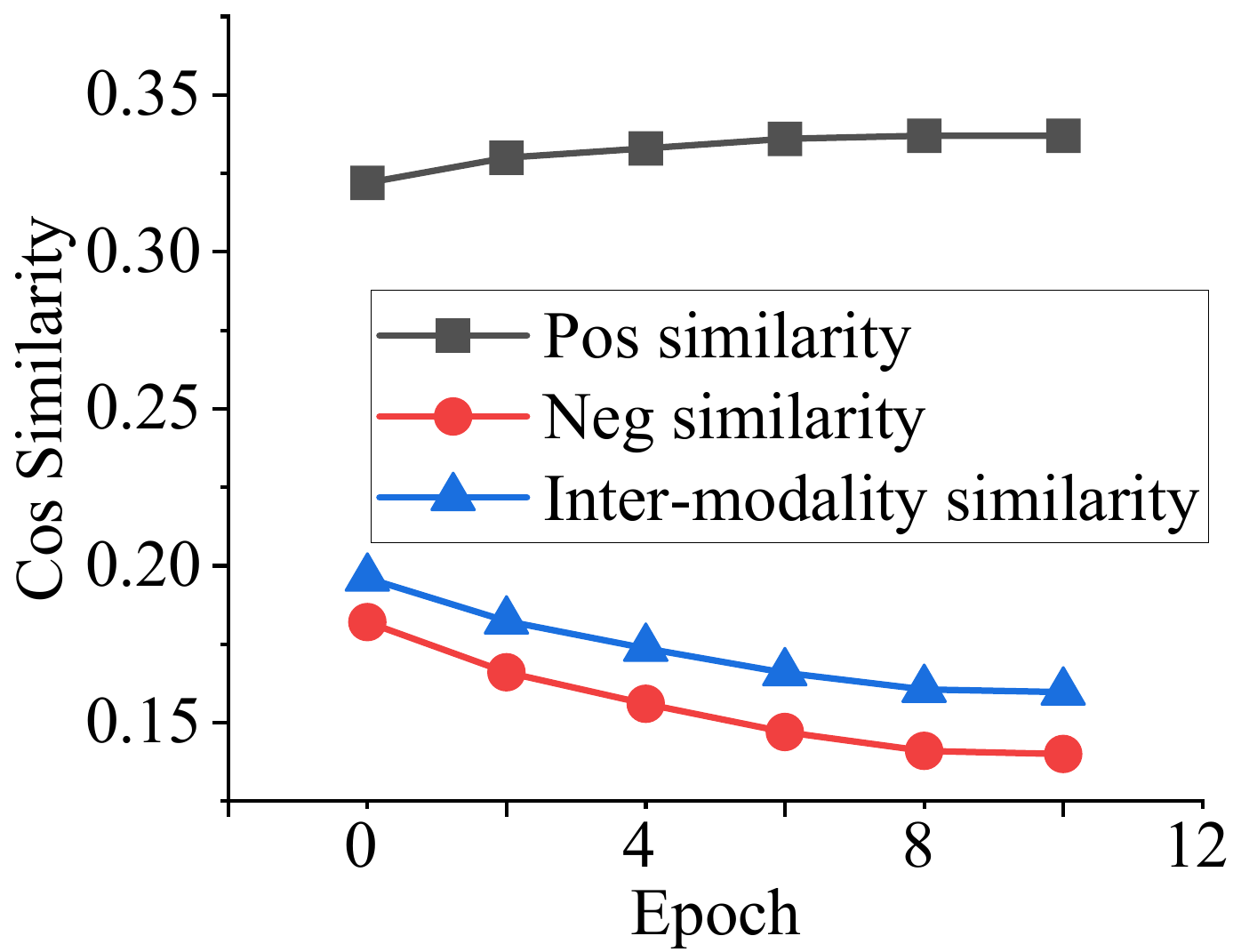}
    \vskip-0.08in
    \caption{Modality gap for a single task changes during training.}
    \label{fig2}
    \includegraphics[width=0.29\textwidth]{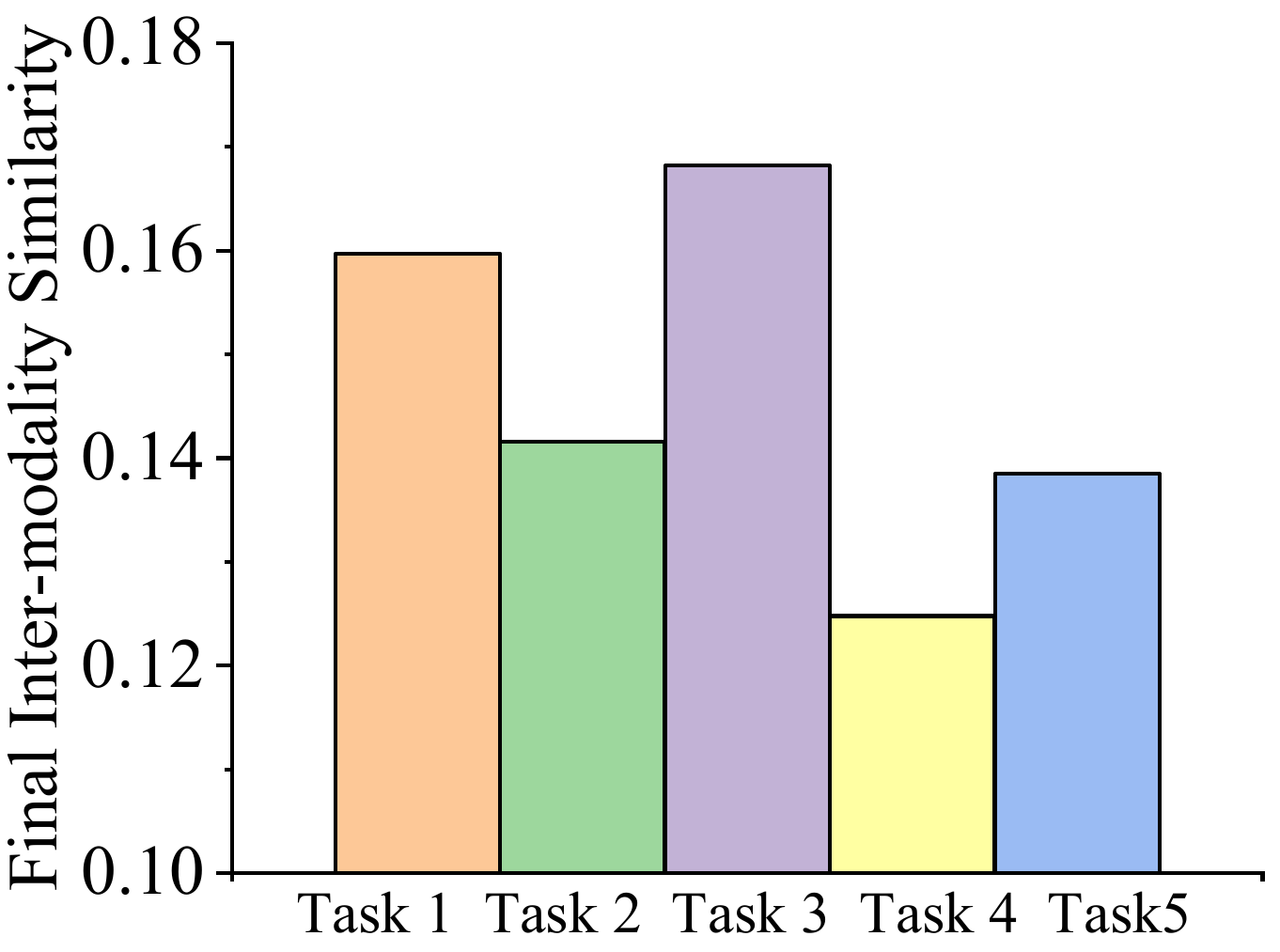}
    \vskip-0.08in
    \caption{The final modality gaps are inconsistent.}
    \label{fig3}
    \vskip-1em
\end{wrapfigure}


We follow prior work~\citep{huang2025mind} to define and measure the modality gap. Given $N$ image features $\{\mathbf{v}_i\}_{i=1}^{N}$ and $\mathcal{C}$ class text features $\{\mathbf{t}_{j}\}_{j=1}^{\mathcal{C}}$ for a task, we can use the average cosine similarity over all image-text pairs to measure the inter-modality similarity, i.e., $\mathrm{sim}=\frac{1}{N}\sum_{i=1}^{N}\frac{1}{\mathcal{C}}\sum_{j=1}^{\mathcal{C}}\langle \mathbf{v}_i,\mathbf{t}_{j}\rangle$. This quantity reflects the closeness between the image and text modalities in the shared feature space. We also compute the average similarity of positive and negative image-text pairs, respectively.

Continual adaptation to downstream tasks perturbs the modality gap. Taking ImageNet-100~\citep{deng2009imagenet} as an example, we measure the change of overall inter-modality similarity during single-task adaptation. In Fig.~\ref{fig2}, as training proceeds, the overall similarity between image and text features gradually decreases. Furthermore, we find that the the similarity of negative pairs drops significantly, dominating the enlargement of the modality gap. Intuitively, this is because downstream training only pulls an image feature closer to its corresponding text feature, while simultaneously pushing it away from a large number of non-matching text features. Prior work has shown that an enlarged modality gap can weaken pre-trained knowledge and limit intra-task discrimination~\citep{huang2025mind}. 

When the model learns task-specific modules, the adaptation process of each task becomes relatively independent. We observe that the modality gaps formed after adaptation are not consistent across tasks, as shown in Fig.~\ref{fig3}. As a result, different task-specific modules may respond unevenly to the same sample during unified inference. For example, if a task ends up with a relatively smaller modality gap, some of its negative image-text pairs may still maintain high similarity scores, thereby interfering with the responses of other tasks. This observation suggests that the model should further compensate modality-gap-induced residual bias so as to enhance intra-task discrimination and enlarge the response margin between the GT task and competing tasks during routing.


\section{Method}
\label{Method}

\begin{figure}[t]
  \centering
  \begin{subfigure}[t]{0.91\linewidth}
    \includegraphics[width=\linewidth]{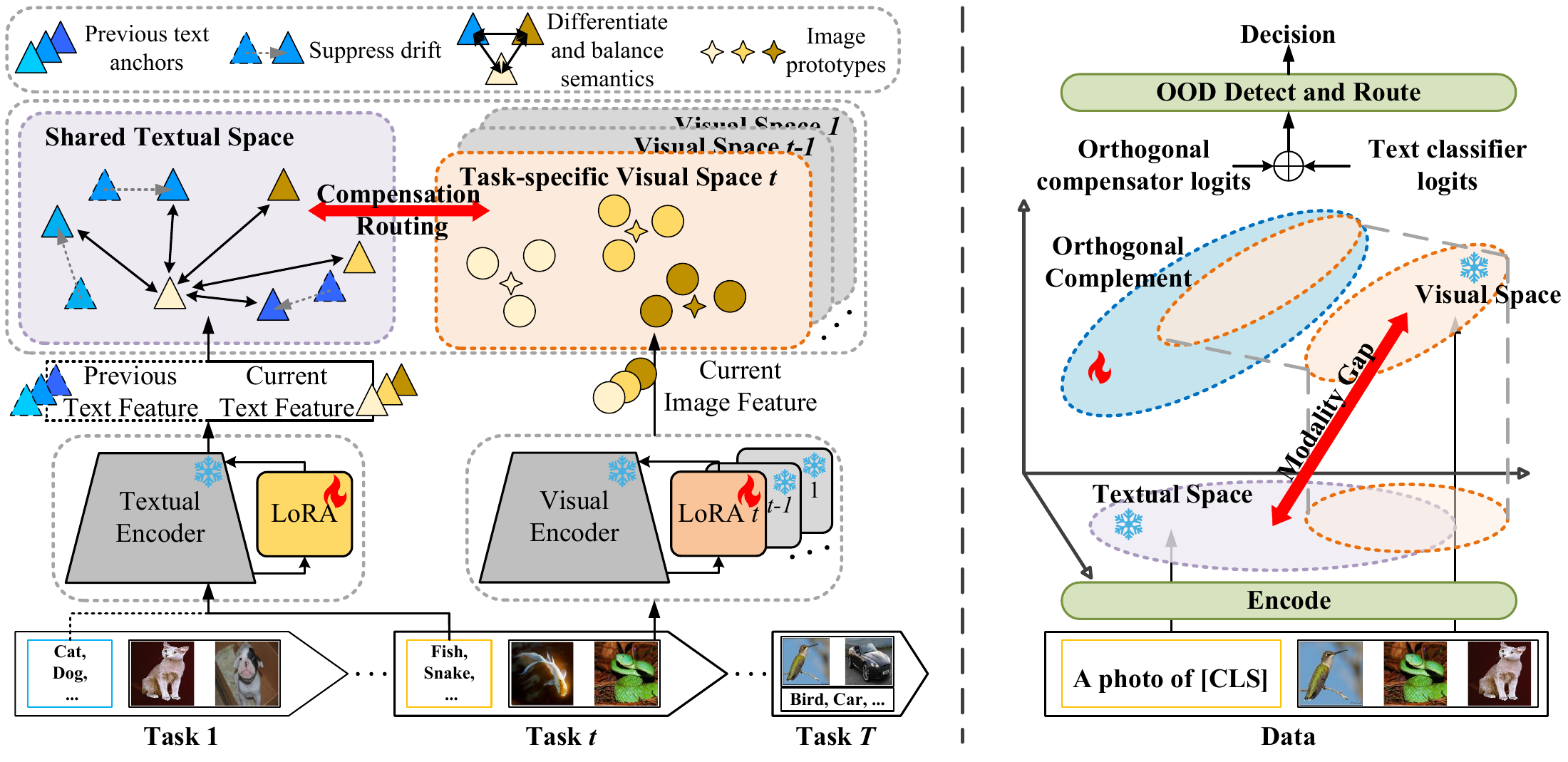}
  \end{subfigure}
  \vskip -0.1in
  \caption{Left: GR4CIL fine-tunes task-specific LoRA modules in the visual branch to accommodate knowledge. In the shared text branch, it anchors the semantics of previous classes while maintaining a unified degree of separation among all text features. Right: To address modality gaps, GR4CIL learns compensation weights in the orthogonal complement of the text space.}
  \vskip -0.22in
  \label{fig4}
\end{figure}

To preserve task-specific knowledge and support reliable routing, we propose GR4CIL. GR4CIL equips the visual branch with task-specific modules, while maintaining a shared and stable semantic space on the text branch (Sec.~\ref{module}). From the perspective of modality gap, we introduce an orthogonal compensation mechanism to enhance intra-task discriminability and make the GT task more separable from competing tasks (Sec.~\ref{compensation}). Based on this, GR4CIL performs task-aware routing with unified score competition and prototype-driven OOD detection (Sec.~\ref{ood}). Fig.~\ref{fig4} illustrates the pipeline.


\subsection{Decoupled Incremental Knowledge Learning}
\label{module}

To simultaneously avoid inter-task knowledge interference and maintain a stable semantic reference across incremental tasks, GR4CIL learns a shared LoRA~\citep{hu2022lora} module on the text branch, while assigning task-specific LoRA modules to the visual branch for each task. These LoRA modules are inserted into the Transformer blocks of the encoders and are applied to the key and value weights. 




We first construct a shared text space, which serves as a stable semantic reference across tasks. Since class prompts remain reusable throughout the incremental process, we explicitly constrain the shared text mapping to preserve the semantics of previously learned classes while keeping newly introduced classes sufficiently separated. Let the current task be the $t$-th task with class set $\mathcal{C}^t$, and the previous classes are $\mathcal{C}^{1:t-1}$. We denote the shared text LoRA at task $t$ by $\phi_{\mathrm{text}}^t$. For each previous class $c \in \mathcal{C}^{1:t-1}$, let $\mathbf{z}_{c}$ be the cached text feature obtained when the class is first learned and retained as a semantic anchor. Meanwhile, let $\mathbf{t}_{c}^{t}=f_t(\mathbf{p}(c);\phi_{\mathrm{text}}^t)$ denote its current normalized text feature. To suppress semantic drift in the shared text space, we define the anchor loss as:
\vskip-0.18in
\begin{equation}
\mathcal{L}_{\mathrm{anc}}=
\frac{1}{|\mathcal{C}^{1:t-1}|}
\sum_{c\in\mathcal{C}^{1:t-1}}
\left(1-\cos(\mathbf{t}_{c}^{t},\mathbf{z}_{c})\right),
\end{equation}
\vskip-0.08in
which encourages the semantic representations of previous classes to remain stable. Meanwhile, we impose a bounded separation constraint only on the new classes. Specifically, for each $c \in \mathcal{C}^t$, we penalize overly high cosine similarity between its text feature and those of the other seen classes $c^\prime$:
\vskip-0.18in
\begin{equation}
\mathcal{L}_{\mathrm{sep}}=
\frac{1}{|\mathcal{C}^{t}|}
\sum_{c\in\mathcal{C}^{t}}
\frac{1}{|\mathcal{C}^{1:t}\setminus\{c\}|}
\sum_{c^\prime\in\mathcal{C}^{1:t}\setminus\{c\}}
\max\left(0,\cos(\mathbf{t}_{c}^{t},\mathbf{t}_{c^\prime}^{t})-\tau\right),
\end{equation}
\vskip-0.08in
where $\tau$ is a fixed separation threshold. The separation loss prevents newly introduced classes from becoming overly close to existing ones, while avoiding unbounded repulsion once sufficient separation is achieved. This not only differentiates text features to enhance discrimination, but also preserves a unified and controllable angular distribution among different classes.

On the visual branch, we assign an independent LoRA module to each task, so that task-specific visual knowledge can be accommodated in a decoupled manner and inter-task interference can be reduced. Let the visual LoRA of task $t$ be denoted by $\phi_{\mathrm{vis}}^t$. When learning task $t$, we freeze all previously learned visual LoRA modules and optimize only the current visual LoRA $\phi_{\mathrm{vis}}^t$ together with the shared text LoRA, updated from $\phi_{\mathrm{text}}^{t-1}$ to $\phi_{\mathrm{text}}^{t}$. Given an input image $\mathbf{x}_i$, its visual feature is denoted by $\mathbf{v}_i^t=f_v(\mathbf{x}_i;\phi_{\mathrm{vis}}^t)$. For each sample $(\mathbf{x}_i,y_i)\in\mathcal{D}^t$, the cosine similarity score for class $c$ is
$s_c(\mathbf{x}_i)=\langle \mathbf{v}_i^t,\mathbf{t}_{c}^{t}\rangle$,
and the downstream adaptation objective is:


\vskip-0.22in
\begin{equation}\label{eq3}
\mathcal{L}_{\mathrm{base}}=
\mathcal{L}_{\mathrm{clip}}+
\lambda_{\mathrm{anc}}\mathcal{L}_{\mathrm{anc}}+
\lambda_{\mathrm{sep}}\mathcal{L}_{\mathrm{sep}}, \qquad
\mathcal{L}_{\mathrm{clip}}=
-\frac{1}{|\mathcal{D}^t|}
\sum_{(\mathbf{x}_i,y_i)\in\mathcal{D}^t}
\log
\frac{\exp(s_{y_i}(\mathbf{x}_i))}
{\sum_{c\in\mathcal{C}^t}\exp(s_c(\mathbf{x}_i))},
\end{equation}
\vskip-0.1in


where $\lambda_{\mathrm{anc}}$ and $\lambda_{\mathrm{sep}}$ are balancing coefficients, and $\mathcal{L}_{\mathrm{clip}}$ is a CLIP-style classification loss. In this way, the shared text branch maintains a stable semantic reference across tasks, while the visual branch preserves task-specific knowledge in a decoupled form.

\subsection{Orthogonal Compensation Based on Modality Gap}
\label{compensation}

In Sec.~\ref{insight}, we show that continual adaptation enlarges the modality gap within each task, while task-specific learning also leads to gap inconsistency across tasks. As a result, although Sec.~\ref{module} establishes a stable text space and decoupled task-specific visual features, classification still relies on text features as classifiers. Due to the modality gap, the discriminative structure of the visual features may not be fully covered by the text space, leaving task-relevant residual directions unused and aggravating the competition among task-specific modules during inference. Motivated by this view, GR4CIL introduces a residual compensation in the orthogonal complement of the text space.

Conventional image-text classification uses text features as classifiers. However, because the textual and visual subspaces only partially overlap~\citep{huang2025mind}, adaptation based solely on the text classifier can be suboptimal. To clarify this point, we consider a simple approximation-based view. For task $t$, let the text feature matrix be
$
\mathbf{T}^t = [\mathbf{t}_{c_1}^t, \mathbf{t}_{c_2}^t, \ldots, \mathbf{t}_{c_{|\mathcal{C}^t|}}^t].
$ We perform SVD on $\mathbf{T}^t$ as $
\mathbf{T}^t = \mathbf{U}_t \mathbf{\Sigma}_t \mathbf{V}_t^\top,
$ and take the column space of $\mathbf{U}_t$ as the text subspace. The projection matrix onto the text subspace is
$
\mathbf{P}_t = \mathbf{U}_t \mathbf{U}_t^\top,
$
while the projection matrix onto its orthogonal complement is
$
\mathbf{P}_t^\perp = \mathbf{I} - \mathbf{U}_t \mathbf{U}_t^\top.
$

\begin{proposition}\label{proposition_1}
Let $W_t^\star$ denote an ideal linear classifier in the full visual feature space, used here only for analysis. If the classifier is constrained to lie in the text subspace, then its best approximation is $W_{t,\mathrm{text}}^\star = \mathbf{P}_t W_t^\star$, and the corresponding approximation error is
$
\mathcal{E}_{\mathrm{text}}^t = \|(\mathbf{I} - \mathbf{P}_t) W_t^\star\|_F^2.
$
\end{proposition}

\begin{lemma}\label{lemma1}
Let $r_t = \mathrm{rank}(\mathbf{T}^t)$ and $\rho_t = \mathrm{rank}(W_t^\star)$. Following \citep{huang2025mind}, let $\sigma_1(W_t^\star) \geq \sigma_2(W_t^\star) \geq \cdots \geq \sigma_{\rho_t}(W_t^\star)$ be the singular values of $W_t^\star$. Then the text-subspace approximation error satisfies
\vskip-0.15in
\begin{equation}
\mathcal{E}_{\mathrm{text}}^t
=
\|(\mathbf{I} - \mathbf{P}_t) W_t^\star\|_F^2
\geq
\sum_{j=r_t+1}^{\rho_t} \sigma_j^2(W_t^\star).
\end{equation}
\vskip-0.1in
Equality holds if and only if the text subspace covers the leading $r_t$ left singular directions.
\end{lemma}

See the Appendix~\ref{proof} for proof. Proposition~\ref{proposition_1} and Lemma~\ref{lemma1} suggest that restricting classification to the text subspace may leave residual discriminative directions uncovered. This motivates learning an additional residual compensation in the orthogonal complement of the text space. In practice, we parameterize this residual term by a task-specific linear head whose column space lies in $\mathrm{span}(\mathbf{P}_t^\perp)$. Specifically, we introduce a compensation head $W_{\mathrm{comp}}^t$ for each task and project it as
$
\widehat{W}_{\mathrm{comp}}^t = \mathbf{P}_t^\perp W_{\mathrm{comp}}^t.
$
For a sample $\mathbf{x}_i$ with visual feature $\mathbf{v}_i^t$, the compensation logits are:
\vskip-0.1in
\begin{equation}
\mathbf{g}^t(\mathbf{x}_i)=\mathbf{v}_i^{t\top}\widehat{W}_{\mathrm{comp}}^t,\qquad
g_c^t(\mathbf{x}_i)=\big[\mathbf{g}^t(\mathbf{x}_i)\big]_c,\qquad
\widehat{s}_c(\mathbf{x}_i)=s_c(\mathbf{x}_i)+\beta\, g_c^t(\mathbf{x}_i),
\end{equation}
\vskip-0.05in
where $g_c^t(\mathbf{x}_i)$ denotes the logit of class $c$ from the compensation head and $\beta$ is a balancing coefficient. Since the compensation head is constrained in the orthogonal complement of the text space, it acts as a residual term and is encouraged to capture directions not covered by the text classifier. The above design can be further justified from the perspective of approximation error in a direct-sum subspace.

\begin{proposition}\label{proposition_2}
Let $\mathcal{S}_R^t \subseteq \mathrm{span}(\mathbf{P}_t^\perp)$ be a compensation subspace, and let its projection matrix be $\mathbf{P}_{R,t}$. Since $\mathrm{span}(\mathbf{P}_t)$ and $\mathcal{S}_R^t$ are orthogonal, the best approximation of $W_t^\star$ in subspace
$
\mathrm{span}(\mathbf{P}_t)\oplus \mathcal{S}_R^t
$
is
$
W_{t,\oplus}^\star = (\mathbf{P}_t + \mathbf{P}_{R,t}) W_t^\star,
$
with approximation error:
$
\mathcal{E}_{\oplus}^t
=
\|(\mathbf{I} - \mathbf{P}_t - \mathbf{P}_{R,t})W_t^\star\|_F^2.
$
\end{proposition}

\begin{corollary}\label{corollary}
Under the condition of Proposition~\ref{proposition_2}, we have
$
\mathcal{E}_{\oplus}^t \leq \mathcal{E}_{\mathrm{text}}^t,
$
and the error reduction is
\vskip-0.1in
\begin{equation}
\mathcal{E}_{\mathrm{text}}^t - \mathcal{E}_{\oplus}^t
=
\|\mathbf{P}_{R,t}(\mathbf{I} - \mathbf{P}_t)W_t^\star\|_F^2.
\end{equation}
\vskip-0.05in
The approximation error is monotonically non-increasing after introducing orthogonal compensation.
\end{corollary}


See the Appendix~\ref{proof} for proof. Corollary~\ref{corollary} further suggests that the reduction in approximation error comes from the residual discriminative energy captured in the orthogonal complement, providing a justification for modeling a residual classifier beyond the text subspace. Such compensation may recover discriminative directions not covered by the text classifier, thereby improving intra-task discrimination. Moreover, since the text subspace is induced by the same shared and stabilized text branch, the corresponding compensations are anchored to a common semantic reference, which makes residual responses across tasks more comparable and benefits score separability during inference.

In practice, this compensation subspace is parameterized by the head $\widehat{W}_{comp}^{t}=\mathbf{P}_t^\perp W_{comp}^{t}$.
Here, $\mathbf{P}_t^\perp$ is computed from the text features of task $t$. Since the compensation space is high-dimensional and the optimization is non-convex, a suitable initialization is beneficial. To better align the compensation head with the visual space, we initialize $W_{\mathrm{comp}}^t$ with the visual class prototypes of the current task. After completing the LoRA learning, we freeze $\phi_{\mathrm{vis}}^t$ and $\phi_{\mathrm{text}}^t$, initialize and optimize the compensation head of task $t$. The compensation head is trained with the cross-entropy objective:
\vskip-0.12in
\begin{equation}\label{eq7}
\mathcal{L}_{\mathrm{comp}}
=
-\frac{1}{|\mathcal{D}^t|}
\sum_{(\mathbf{x}_i,y_i)\in\mathcal{D}^t}
\log
\frac{\exp(g_{y_i}^t(\mathbf{x}_i))}
{\sum_{c\in\mathcal{C}^t}\exp(g_{c}^t(\mathbf{x}_i))}.
\end{equation}
\vskip-0.1in
See the Appendix~\ref{Algorithms} for the pseudocode. By learning a modality-gap-guided compensation classifier, GR4CIL compensates for the discriminative deficiency of the text classifier, while providing a more favorable score basis for separating the GT task from competing tasks during task-aware routing.



\subsection{Routing and Inference}
\label{ood}


Regarding CIL, after learning task-specific knowledge and compensation, GR4CIL performs inference with unified score competition and prototype-based task awareness. For each class $c\in\mathcal{C}^t$, we maintain a visual class prototype $\mathbf{p}_c^t$ and use its similarity to the sample feature as an additional OOD cue. Accordingly, for a sample $\mathbf{x}_i$, the final score of class $c$ is defined as
$
q_c(\mathbf{x}_i)=\widehat{s}_c(\mathbf{x}_i)+\gamma\,\langle \mathbf{v}_i^t,\mathbf{p}_c^t\rangle,
$
where $\gamma$ is a fixed hyper-parameter. This prototype-based OOD cue makes the decision depend not only on text matching and compensation, but also on the consistency between the sample and the class distribution structure. The final prediction is then obtained by
$
\widehat{y}_i=\arg\max_{c\in\mathcal{C}^{1:t}} q_c(\mathbf{x}_i).
$
In this way, routing is realized through score competition among classes from different task-specific branches. The above scoring rule defines the standard unified inference over all seen classes.

Beyond CIL setting, we leave an unknown-aware interface that can be used to extend inference toward zero-shot generalization. We equip each task with an acceptance threshold $\omega^t$. For task $t$, we compute the softmax confidence within its own class set $\mathcal{C}^t$ and define the task-level maximum softmax probability (MSP) $m_t(\mathbf{x}_i)$ over the scores $\{q_c(\mathbf{x}_i)\mid c\in\mathcal{C}^t\}$. Task $t$ accepts the sample if $m_t(\mathbf{x}_i)>\omega^t$. A sample is regarded as potentially unknown only when it is rejected by all tasks. In practice, $\omega^t$ can be estimated from validation data without using future-task or OOD samples. This unknown-aware branch provides a simple interface for extending inference to more open scenarios.

For a potentially unknown sample $\mathbf{x}_i^*$, GR4CIL can optionally leverage the learned knowledge for generalized prediction. Specifically, we extract task-conditioned visual features $\mathbf{v}_i^{t*}$ from all task-specific branches and define the confidence of task $t$ as
$
r_t(\mathbf{x}_i^*)=\max_{c\in\mathcal{C}^t}\langle \mathbf{v}_i^{t*},\mathbf{p}_c^t\rangle.
$
We convert these confidences into task weights and perform zero-shot classification over a candidate label set:
\vskip-0.2in
\begin{equation}
q_{k}^{\text{fuse}}(\mathbf{x}_i^*)=\sum_{\tau=1}^{T}\alpha_\tau(\mathbf{x}_i^*)\langle \mathbf{v}_i^{\tau *},\mathbf{t}_{k}\rangle,\quad
\alpha_t(\mathbf{x}_i^*)=
\frac{\exp(r_t(\mathbf{x}_i^*))}
{\sum_{\tau=1}^{T}\exp(r_\tau(\mathbf{x}_i^*))},
\end{equation}
\vskip-0.08in

where $\mathbf{t}_{k}$ denotes the text feature of zero-shot class $k$ encoded by the shared text branch. In this way, while the standard CIL prediction is still given by the unified routing rule above, GR4CIL also leaves a practical interface for confidence-based extension beyond the learned label space. In the worst case, the model can fall back to the original CLIP for broader zero-shot inference.

\section{Experiments}
\label{Experiments}

\subsection{Experimental Setup}

\begin{table*}[t]
\centering
\small
\setlength{\tabcolsep}{4pt}
\renewcommand{\arraystretch}{1.12}
\caption{Comparison of CIL performance on four benchmarks. All datasets are split into 10 tasks. `Avg' and `Last' denote the average accuracy (Avg-Acc) and the final accuracy (Last-Acc), respectively. For GR4CIL, we report the mean and standard deviation over three independent runs.}
\vskip-0.08in
\begin{tabular}{l cc cc cc cc cc}
\toprule
\multirow{2}{*}{Method}
& \multicolumn{2}{c}{CIFAR-100}
& \multicolumn{2}{c}{ImageNet-R}
& \multicolumn{2}{c}{ImageNet100}
& \multicolumn{2}{c}{ImageNet-1K}
& \multicolumn{2}{c}{Average} \\
\cmidrule(lr){2-3} \cmidrule(lr){4-5} \cmidrule(lr){6-7} \cmidrule(lr){8-9} \cmidrule(lr){10-11}
& Avg & Last & Avg & Last & Avg & Last & Avg & Last & Avg-Acc & Last-Acc \\
\midrule

L2P++          
& 81.90 & 73.08 & 81.67 & 75.98 & 80.51 & 67.22 & 79.30 & 69.60 & 80.84 & 71.47 \\
DualPrompt     
& 81.45 & 72.51 & 82.01 & 75.77 & 80.65 & 67.38 & 79.39 & 69.79 & 80.88 & 71.36 \\
CODA           
& 76.98 & 62.25 & 78.00 & 67.52 & 64.13 & 34.76 & 76.99 & 66.96 & 74.03 & 57.87 \\
Aper-Adapter   
& 75.76 & 63.50 & 78.65 & 71.35 & 85.84 & 76.40 & 76.60 & 68.74 & 79.21 & 70.00 \\
Continual-CLIP 
& 75.15 & 66.68 & 79.12 & 72.00 & 84.98 & 75.40 & 72.96 & 64.44 & 78.05 & 69.63 \\
CLAP           
& 74.19 & 63.45 & 81.22 & 75.80 & 81.07 & 72.00 & 75.85 & 67.36 & 78.08 & 69.65 \\
MOE4CL         
& 85.36 & 78.37 & 85.28 & 80.77 & 86.39 & 76.66 & 81.29 & 72.73 & 84.58 & 77.13 \\
MagMax         
& 85.63 & 79.00 & 87.13 & 80.85 & 86.33 & 75.92 & 80.74 & 71.31 & 84.96 & 76.77 \\
MG-CLIP        
& 87.00 & 80.57 & 87.58 & \textit{82.67} & \textit{87.31} & \textit{78.38} & 81.88 & \textit{73.68} & 85.94 & \textit{78.83} \\
AdapterVLM     
& \textit{87.98} & \textit{81.65} & \textit{88.25} & 82.51 & 86.03 & 77.05 & \textit{81.60} & 73.47 & \textit{85.97} & 78.67 \\
\midrule

\multirow{2}{*}{GR4CIL (Ours)}
& \textbf{89.35} & \textbf{83.22} & \textbf{89.50} & \textbf{84.15}
& \textbf{87.65} & \textbf{78.64} & \textbf{83.36} & \textbf{75.33}
& \textbf{87.47} & \textbf{80.34} \\
& {\scriptsize $\pm$0.26} & {\scriptsize $\pm$0.28}
& {\scriptsize $\pm$0.16} & {\scriptsize $\pm$0.11}
& {\scriptsize $\pm$0.29} & {\scriptsize $\pm$0.33}
& {\scriptsize $\pm$0.30} & {\scriptsize $\pm$0.31}
& {\scriptsize \textbf{(+1.50)}} & {\scriptsize \textbf{(+1.51)}} \\
\bottomrule
\end{tabular}
\vskip-0.2in
\label{tab:main_results}
\end{table*}

\textbf{Datasets.} We evaluate our method on four commonly used CIL benchmarks, including CIFAR-100~\citep{CIFAR-100}, ImageNet-R~\citep{hendrycks2021many}, ImageNet100~\citep{deng2009imagenet}, and ImageNet-1K~\citep{deng2009imagenet}. All datasets are evenly divided into 10 sequential tasks under the standard CIL setting. For the comparison under different task sequence lengths, we further construct additional task partitions with 5-task and 20-task settings.


\textbf{Baselines.} To ensure a fair comparison, we focus exclusively on replay-free CIL methods~\citep{zhou2022model}. Specifically, we compare GR4CIL with representative CLIP-based methods, including Continual-CLIP~\citep{thengane2022clip}, CLAP~\citep{jha2024clap4clip}, MOE4CL~\citep{yu2024boosting}, MagMax~\citep{marczak2024magmax}, MG-CLIP~\citep{huang2025mind}, and AdapterVLM~\citep{zhang2025visual}. We further include several advanced vision-only methods, including L2P++~\citep{wang2022learning}, DualPrompt~\citep{wang2022dualprompt}, CODA~\citep{smith2023coda}, and Aper-Adapter~\citep{zhou2025revisiting}. All methods adopt the ViT-B/16 weights of OpenAI~\citep{radford2021learning} by default. Baseline results are taken from prior papers~\citep{huang2025mind, zhang2025visual} or reproduced using their publicly available code.


\textbf{Evaluation metrics.} We adopt average accuracy (Avg-Acc) and final accuracy (Last-Acc) as the primary evaluation metrics for CIL. Let $A_t$ denote the classification accuracy after learning the $t$-th task and evaluating on all seen classes up to task $t$. Then, Avg-Acc is defined as the average accuracy, i.e.,
$
\frac{1}{T}\sum_{t=1}^{T} A_t,
$
while Last-Acc denotes the final test accuracy, i.e., $A_T$. To evaluate the OOD detection capability of GR4CIL, we further report Avg-AUROC and Last-AUROC~\citep{zhang2025visual}. Specifically, at the $t$-th incremental stage, all previously learned classes are regarded as in-distribution (ID) classes, while the classes that have not yet been learned in future tasks are treated as OOD classes. For each test sample, we compute the MSP based on the model output scores and use it for OOD detection~\citep{hendrycks2016baseline}. The AUROC is then computed at each incremental stage, and we finally summarize the results as the average AUROC over all stages (Avg-AUROC) and the AUROC at the final stage (Last-AUROC).

\textbf{Implementation details.} We follow the basic training protocol of prior work~\citep{zhang2025visual}. We adopt CLIP ViT-B/16 as the pretrained backbone. Balancing coefficients $\lambda_{\mathrm{anc}}$ and $\lambda_{\mathrm{sep}}$ are set to 1. We use AdamW as the optimizer with a learning rate of 0.005, which is scheduled by cosine annealing. The LoRA rank is set to 24 for both textual and visual branch. The separation threshold $\tau$ is set to 0.7. The compensation head is trained separately for each task using Adam with a learning rate of 0.0005. During inference, the coefficients $\beta$ and $\gamma$ are both set to 0.2. For more details, see the Appendix~\ref{Implementation}.


\subsection{Comparison Results}

In Table~\ref{tab:main_results}, it can be seen that GR4CIL consistently outperforms existing replay-free methods across the four benchmark datasets. Specifically, on CIFAR-100, GR4CIL achieves 89.35\% Avg-Acc and 83.22\% Last-Acc, surpassing the strongest baseline by 1.37\% and 1.57\%. On ImageNet-R, it further improves over the best baseline by 1.25\% in Avg-Acc and 1.48\% in Last-Acc. On the more challenging ImageNet100 and large-scale ImageNet-1K, GR4CIL continues to achieve the best results. These results demonstrate that GR4CIL can still learn more robust task-specific knowledge and maintain stronger performance as new classes are continuously accumulated. To further evaluate the robustness of our method under different incremental granularities, we conduct additional experiments with different task sequence lengths, as reported in Table~\ref{tab:seq_and_ood} (left). GR4CIL achieves the best results under both the 5-task and 20-task settings on CIFAR-100, as well as under the 5-task setting on ImageNet-R, while remaining competitive under the 20-task setting on ImageNet-R. These results suggest that GR4CIL generalizes well across different incremental sequences.

We further evaluate the OOD detection capability of GR4CIL in Table~\ref{tab:seq_and_ood} (right) to examine the feasibility of extending the standard CIL inference beyond the learned label space. In this setting, the samples from all learned classes at the current stage are regarded as ID samples, while the samples belonging to future unseen tasks are treated as OOD samples. The results show that GR4CIL consistently outperforms existing methods in terms of Avg-AUROC and Last-AUROC, indicating that the learned representations capture whether a sample lies inside or outside the currently acquired knowledge scope. Therefore, these results support the feasibility of using the proposed confidence signals as a trigger for extending inference beyond the standard CIL setting.

\begin{table*}[t]
\centering
\small
\setlength{\tabcolsep}{4pt}
\renewcommand{\arraystretch}{1.08}
\caption{Left: Comparison of final accuracy (Last-Acc) under different task sequence lengths on CIFAR-100 and ImageNet-R. Right: Comparison of OOD detection performance, where Avg-AUROC and Last-AUROC are used as evaluation metrics and the task sequence length is set to 10.}
\vskip-0.08in
\begin{subtable}[t]{0.485\textwidth}
\centering
\label{tab:seq_length}
\begin{tabular}{lcccc}
\toprule
Method & \multicolumn{2}{c}{CIFAR-100} & \multicolumn{2}{c}{ImageNet-R} \\
\cmidrule(lr){2-3} \cmidrule(lr){4-5}
& 5-task & 20-task & 5-task & 20-task \\
\midrule
MOE4CL   & 78.96 & 76.20 & 81.37 & 79.58 \\
MagMax   & 82.07 & 76.84 & 82.75 & 80.18 \\
MG-CLIP  & 81.47 & 79.31 & 83.13 & 82.12 \\
AdapterVLM   & 83.77 & 77.62 & 83.20 & 80.45 \\
\midrule
Ours     & \textbf{84.98} & \textbf{79.52} & \textbf{85.32} & \textit{81.36} \\
\bottomrule
\end{tabular}
\end{subtable}
\hfill
\begin{subtable}[t]{0.485\textwidth}
\centering
\label{tab:ood_detection}
\begin{tabular}{lcccc}
\toprule
Method & \multicolumn{2}{c}{CIFAR-100} & \multicolumn{2}{c}{ImageNet-R} \\
\cmidrule(lr){2-3} \cmidrule(lr){4-5}
& Avg & Last & Avg & Last \\
\midrule
CODA & 80.97 & 76.36 & 81.04 & 75.53 \\
Continual-CLIP & 78.43 & 72.28 & 80.97 & 78.10 \\
MOE4CL & 80.12 & 74.76 & 82.32 & 76.73 \\
AdapterVLM & 88.59 & 85.89 & 88.35 & 82.90 \\
\midrule
Ours       & \textbf{89.13} & \textbf{87.15} & \textbf{88.75} & \textbf{83.09} \\
\bottomrule
\end{tabular}
\end{subtable}
\vskip-0.1in

\label{tab:seq_and_ood}
\end{table*}

\begin{table*}[t]
\centering
\small
\setlength{\tabcolsep}{5pt}
\renewcommand{\arraystretch}{1.10}

\caption{Left: Zero-shot generalization on unseen datasets. GR4CIL i uses CIFAR-100 for training, and GR4CIL ii uses ImageNet-R. Right: Distance between the image space and classifier spaces.}
\vskip-0.08in
\begin{subtable}[t]{0.48\textwidth}
\centering
\label{tab:zs_generalization}
\begin{tabular}{lccc}
\toprule
Method & PETS & FOOD101 & ImageNet-1K \\
\midrule
CLIP & 84.96 & 83.49 & 65.41 \\
GR4CIL i & 86.23 & 83.76 & 66.52 \\
GR4CIL ii & 86.72 & 83.52 & 66.87 \\
\bottomrule
\end{tabular}
\end{subtable}
\hfill
\begin{subtable}[t]{0.51\textwidth}
\centering
\label{tab:subspace_distance}
\begin{tabular}{lccc}
\toprule
 & CIFAR100 & ImageNet-R & ImageNet100 \\
\midrule
I-T   & 0.814 & 0.767 & 0.811 \\
I-C   & 0.612 & 0.670 & 0.722 \\
I-TC  & 0.224 & 0.210 & 0.420 \\
\bottomrule
\end{tabular}
\end{subtable}

\vskip-0.25in
\label{tab:analysis_tables}
\end{table*}

\subsection{Further Analysis}

Given such a trigger, we next examine whether GR4CIL preserves the zero-shot generalization ability and whether the proposed knowledge fusion can further benefit zero-shot classification on unseen datasets. Specifically, after performing CIL on CIFAR-100 and ImageNet-R, we apply the proposed knowledge fusion to zero-shot classification on Pets~\citep{parkhi2012cats}, Food101~\citep{bossard2014food}, and ImageNet-1K. As shown in Table~\ref{tab:analysis_tables} (left), GR4CIL achieves results comparable to or slightly better than those of the original CLIP on multiple unseen datasets, suggesting that GR4CIL can preserve the generalization capability and the proposed fusion interface can sometimes provide additional benefit. Here, the CLIP results correspond to directly using the original pre-trained CLIP, which also serves as the fallback option in our framework. Since different downstream training data may affect zero-shot behavior, GR4CIL can invoke either the fused prediction or, in the worst case, directly fall back to the original CLIP. It shows that GR4CIL leaves a practical interface for extending inference toward more open scenarios.



We further examine whether orthogonal compensation helps reduce the discrepancy between the classifier space and the image space. To this end, we measure the distances between the image space and three classifier spaces, namely, the text classifier space (I-T), the compensation classifier space (I-C), and their joint space (I-TC), as reported in Table~\ref{tab:analysis_tables} (right). Specifically, we extract the orthonormal bases of the corresponding subspaces from the image features and classifier weights, and use the mean projection residual of the image-space basis vectors onto each target space as the distance measure, where a smaller value indicates better alignment. The results show that the original text classifier space deviates noticeably from the image space, while the joint space after introducing compensation becomes substantially closer. This observation suggests that orthogonal compensation complements discriminative directions not covered by the text classifier, which is consistent with its role in reducing modality-gap-induced mismatch and improving intra-task discriminability.

\begin{figure*}[t]
\centering
\setlength{\tabcolsep}{4pt}
\renewcommand{\arraystretch}{1.05}

\begin{subfigure}[t]{0.48\textwidth}
    \centering
    \includegraphics[width=\linewidth]{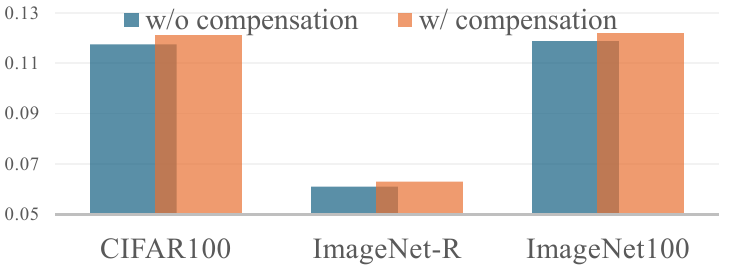}
    \vskip-0.08in
    \caption{Intra-task confidence w/ and w/o. compensation.}
    \label{fig:confidence_bar}
\end{subfigure}
\hfill
\begin{subfigure}[t]{0.48\textwidth}
    \centering
    \includegraphics[width=\linewidth]{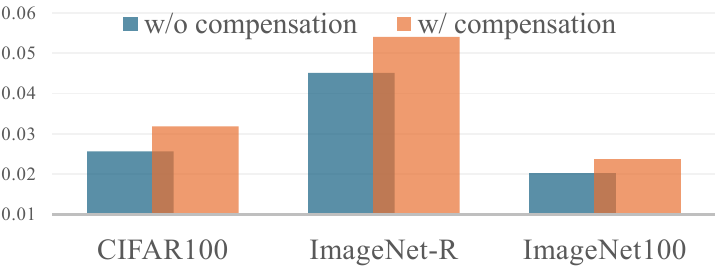}
    \vskip-0.08in
    \caption{Inter-task margin w/ and w/o. compensation.}
    \label{fig:margin_bar}
\end{subfigure}
\hfill

\vskip-0.1in
\caption{Effect of orthogonal compensation on prediction confidence and inter-task margin.}
\vskip-0.2in
\label{fig:compensation_analysis}
\end{figure*}

Beyond this geometric analysis, we study how orthogonal compensation affects prediction confidence and cross-task score separation. After introducing compensation, the average intra-task softmax confidence of the GT class consistently increases (see Fig.~\ref{fig:compensation_analysis}(a)). We also measure the margin between the score of the GT task and the highest competing score from other tasks, and find that this margin is consistently enlarged after compensation, with an average improvement of 20.34\% (see Fig.~\ref{fig:compensation_analysis}(b)). Correspondingly, the average routing accuracy across the three datasets improves from 81.37\% to 82.85\%. These results indicate that compensation not only improves intra-task discrimination, but also leads to clearer score separation among competing tasks, supporting more reliable routing.

\begin{table}[t]
\centering
\small
\setlength{\tabcolsep}{5pt}
\renewcommand{\arraystretch}{1.10}
\caption{Ablation study of the main components of GR4CIL on CIFAR-100.}
\label{tab:ablation}
\begin{tabular}{lcccc}
\toprule
Method & Avg-Acc & Last-Acc & Avg-AUROC & Last-AUROC \\
\midrule
Base & 87.17 & 79.40 & 86.46 & 83.33 \\
+ \(\mathcal{L}_{\mathrm{anc}}\) and \(\mathcal{L}_{\mathrm{sep}}\) & 88.74 & 80.88 & 88.30 & 85.82 \\
+ Compensation term & 89.13 & 82.88 & 88.85 & 86.64 \\
+ Prototype term (Full model) & \textbf{89.35} & \textbf{83.22} & \textbf{89.13} & \textbf{87.15} \\
\bottomrule
\end{tabular}
\vskip-0.2in
\end{table}

\begin{wrapfigure}[18]{r}{0.32\textwidth}
    \centering
    \vspace{-1em}

    \begin{minipage}[t]{1.0\linewidth}
        \centering
        \small
        \setlength{\tabcolsep}{3pt}
        \captionof{table}{Analysis of the compensation on CIFAR-100.}
        \vskip-0.15in
        \label{tab:comp_analysis}
        \vspace{0.3em}
        \begin{tabular}{cccc}
            \toprule
            Orth & Proto & Avg-Acc & Last-Acc \\
            \midrule
            $\times$     & $\times$     & 86.68 & 78.64 \\
            $\checkmark$ & $\times$     & 87.34 & 79.22 \\
            $\times$     & $\checkmark$ & 87.95 & 80.52 \\
            $\checkmark$ & $\checkmark$ & 89.35 & 83.22 \\
            \bottomrule
        \end{tabular}
    \end{minipage}

    \begin{minipage}[t]{1.0\linewidth}
        \centering
        \includegraphics[width=\linewidth]{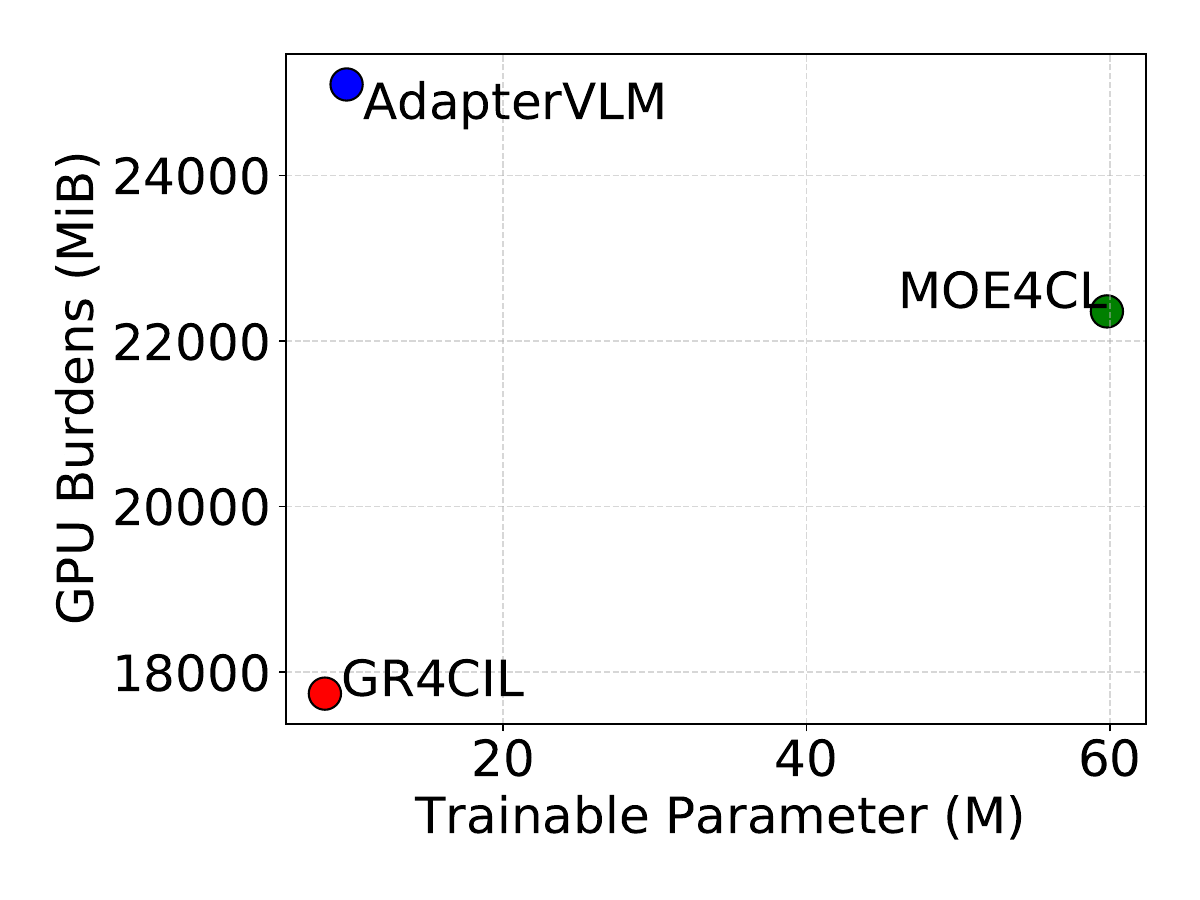}
        \vskip-0.15in
        \captionof{figure}{Computational Cost.}
        \label{fig:cost}
    \end{minipage}\hfill

    \vspace{-1em}
\end{wrapfigure}

Finally, we conduct ablation and component analysis. As shown in Table~\ref{tab:ablation}, $\mathcal{L}_{\mathrm{anc}}$, $\mathcal{L}_{\mathrm{sep}}$, the compensation term, and the prototype term all contribute positively to the final performance. We further analyze the design of the compensation module in Table~\ref{tab:comp_analysis}, where ``Orth'' indicates whether the compensation head is constrained in the orthogonal complement of the text space, and ``Proto'' indicates whether it is initialized with the visual class prototypes. The results show that the orthogonal version consistently outperforms the unconstrained counterpart, suggesting that it is beneficial to model residual discriminative directions beyond the text classifier rather than simply increasing classifier flexibility. Moreover, prototype-based initialization further improves performance in both cases, indicating that initializing the compensation head near the visual structure leads to more effective optimization. We also compare the parameter and memory efficiency in Fig.~\ref{fig:cost}. Specifically, we report the total number of trainable parameters throughout the incremental process and the peak GPU memory usage over both training and inference. The results show that GR4CIL achieves competitive performance with fewer trainable parameters and lower GPU memory burden than other task-specific baselines, indicating a favorable trade-off between effectiveness and efficiency. More detailed analyses of GR4CIL and parameter sensitivity are provided in the Appendix~\ref{routing},~\ref{Semantic} and~\ref{Parameter}.


\section{Conclusion}
\label{Conclusion}


This paper presents GR4CIL for CIL. The proposed method combines task-specific knowledge preservation, modality-gap compensation, and task-aware routing to reduce inter-task interference and improve unified inference over learned classes. In particular, the proposed compensation mechanism recovers residual discriminative directions beyond the text subspace, leading to better intra-task discrimination and clearer separation among competing task-specific branches. In addition, GR4CIL leaves a practical interface for extending inference beyond the standard CIL setting. Future work may further explore scenarios with ambiguous task boundaries and emerging new classes.
\newpage

\bibliographystyle{ieeetr}
\bibliography{main}

@article{french1999catastrophic,
  title={Catastrophic forgetting in connectionist networks},
  author={French, Robert M},
  journal={Trends in Cognitive Sciences},
  volume={3},
  number={4},
  pages={128--135},
  year={1999},
  publisher={Elsevier}
}

@inproceedings{wang2022learning,
  title={Learning to prompt for continual learning},
  author={Wang, Zifeng and Zhang, Zizhao and Lee, Chen-Yu and Zhang, Han and Sun, Ruoxi and Ren, Xiaoqi and Su, Guolong and Perot, Vincent and Dy, Jennifer and Pfister, Tomas},
  booktitle={Proceedings of the IEEE/CVF Conference on Computer Vision and Pattern Recognition},
  pages={139--149},
  year={2022}
}

@inproceedings{wang2022dualprompt,
  title={Dualprompt: Complementary prompting for rehearsal-free continual learning},
  author={Wang, Zifeng and Zhang, Zizhao and Ebrahimi, Sayna and Sun, Ruoxi and Zhang, Han and Lee, Chen-Yu and Ren, Xiaoqi and Su, Guolong and Perot, Vincent and Dy, Jennifer and others},
  booktitle={Computer Vision--ECCV 2022: 17th European Conference, Tel Aviv, Israel, October 23--27, 2022, Proceedings, Part XXVI},
  pages={631--648},
  year={2022},
  organization={Springer}
}

@inproceedings{zhou2022model, 
  title={A model or 603 exemplars: Towards memory-efficient class-incremental learning},
  author={Zhou, Da-Wei and Wang, Qi-Wei and Ye, Han-Jia and Zhan, De-Chuan},
  booktitle={International Conference on Learning Representations},
  pages={},
  year={2023} 
}

@inproceedings{smith2023coda,
  title={CODA-Prompt: COntinual Decomposed Attention-based Prompting for Rehearsal-Free Continual Learning},
  author={Smith, James Seale and Karlinsky, Leonid and Gutta, Vyshnavi and Cascante-Bonilla, Paola and Kim, Donghyun and Arbelle, Assaf and Panda, Rameswar and Feris, Rogerio and Kira, Zsolt},
  booktitle={Proceedings of the IEEE/CVF Conference on Computer Vision and Pattern Recognition},
  pages={11909--11919},
  year={2023}
}

@inproceedings{hendrycks2021many,
  title={The many faces of robustness: A critical analysis of out-of-distribution generalization},
  author={Hendrycks, Dan and Basart, Steven and Mu, Norman and Kadavath, Saurav and Wang, Frank and Dorundo, Evan and Desai, Rahul and Zhu, Tyler and Parajuli, Samyak and Guo, Mike and others},
  booktitle={Proceedings of the IEEE/CVF International Conference on Computer Vision},
  pages={8340--8349},
  year={2021}
}

@inproceedings{li2024CLDNet,
  title={Towards Continual Learning Desiderata via HSIC-Bottleneck Orthogonalization and Equiangular Embedding},
  author={Li, Depeng and Wang, Tianqi and Chen, Junwei and Ren, Qining and Kawaguchi, Kenji and Zeng, Zhigang},
  booktitle={Proceedings of the AAAI Conference on Artificial Intelligence},
  pages={13464--13473},
  year={2024}
}

@article{kim2022theoretical,
  title={A theoretical study on solving continual learning},
  author={Kim, Gyuhak and Xiao, Changnan and Konishi, Tatsuya and Ke, Zixuan and Liu, Bing},
  journal={Advances in Neural Information Processing Systems},
  volume={35},
  pages={5065--5079},
  year={2022}
}

@inproceedings{ming2023cider,
  title={How to Exploit Hyperspherical Embeddings for Out-of-Distribution Detection?},
  author={Yifei Ming and Yiyou Sun and Ousmane Dia and Yixuan Li},
  booktitle={International Conference on Learning Representations },
  pages={},
  year={2023},
}

@inproceedings{lin2024TPL,
      title={Class Incremental Learning via Likelihood Ratio Based Task Prediction}, 
      author={Haowei Lin and Yijia Shao and Weinan Qian and Ningxin Pan and Yiduo Guo and Bing Liu},
      pages={},
      year={2024},
      booktitle={International Conference on Learning Representations}
}

@inproceedings{kim2022MORE,
	title={A multi-head model for continual learning via out-of-distribution replay},
	author={Kim, Gyuhak and Liu, Bing and Ke, Zixuan},
	booktitle={Conference on Lifelong Learning Agents},
	pages={548--563},
	year={2022},
	organization={PMLR}
}

@inproceedings{morteza2022provable,
  title={Provable guarantees for understanding out-of-distribution detection},
  author={Morteza, Peyman and Li, Yixuan},
  booktitle={Proceedings of the AAAI Conference on Artificial Intelligence},
  pages={7831--7840},
  year={2022}
}

@inproceedings{Lu2024PALM,
    title={Learning with Mixture of Prototypes for Out-of-Distribution Detection},
    author={Haodong Lu and Dong Gong and Shuo Wang and Jason Xue and Lina Yao and Kristen Moore},
    booktitle={International Conference on Learning Representations},
    pages={},
    year={2024}
}

@InProceedings{Kim2022CLOM,
    author    = {Kim, Gyuhak and Esmaeilpour, Sepideh and Xiao, Changnan and Liu, Bing},
    title     = {Continual Learning Based on OOD Detection and Task Masking},
    booktitle = {Proceedings of the IEEE/CVF Conference on Computer Vision and Pattern Recognition Workshops},
    year      = {2022},
    pages     = {3856--3866}
}

@inproceedings{li2024harnessing,
author = {Li, Depeng and Wang, Tianqi and Chen, Junwei and Dai, Wei and Zeng, Zhigang},
title = {Harnessing neural unit dynamics for effective and scalable class-incremental learning},
booktitle={International Conference on Machine Learning},
pages={28688--28705},
year={2024}
}

@article{CIFAR-100, 
  title={Learning multiple layers of features from tiny images},
  author={Krizhevsky, Alex and Hinton, Geoffrey and others},
  journal={Handbook of Systemic Autoimmune Diseases},
  year={2009},
  publisher={Citeseer}
}

@inproceedings{deng2009imagenet,  
  title={Imagenet: A large-scale hierarchical image database},
  author={Deng, Jia and Dong, Wei and Socher, Richard and Li, Li-Jia and Li, Kai and Fei-Fei, Li},
  booktitle={2009 IEEE Conference on Computer Vision and Pattern Recognition},
  pages={248--255},
  year={2009},
  organization={IEEE}
}

@article{jha2024clap4clip,
  title={Clap4clip: Continual learning with probabilistic finetuning for vision-language models},
  author={Jha, Saurav and Gong, Dong and Yao, Lina},
  journal={Advances in neural information processing systems},
  volume={37},
  pages={129146--129186},
  year={2024}
}

@inproceedings{radford2021learning,
  title={Learning transferable visual models from natural language supervision},
  author={Radford, Alec and Kim, Jong Wook and Hallacy, Chris and Ramesh, Aditya and Goh, Gabriel and Agarwal, Sandhini and Sastry, Girish and Askell, Amanda and Mishkin, Pamela and Clark, Jack and others},
  booktitle={International conference on machine learning},
  pages={8748--8763},
  year={2021},
  organization={PmLR}
}

@inproceedings{huang2025mind,
  title={Mind the gap: Preserving and compensating for the modality gap in clip-based continual learning},
  author={Huang, Linlan and Cao, Xusheng and Lu, Haori and Meng, Yifan and Yang, Fei and Liu, Xialei},
  booktitle={Proceedings of the IEEE/CVF International Conference on Computer Vision},
  pages={3777--3786},
  year={2025}
}

@inproceedings{yu2024boosting,
  title={Boosting continual learning of vision-language models via mixture-of-experts adapters},
  author={Yu, Jiazuo and Zhuge, Yunzhi and Zhang, Lu and Hu, Ping and Wang, Dong and Lu, Huchuan and He, You},
  booktitle={Proceedings of the IEEE/CVF Conference on Computer Vision and Pattern Recognition},
  pages={23219--23230},
  year={2024}
}

@inproceedings{zhou2025external,
  title={External knowledge injection for clip-based class-incremental learning},
  author={Zhou, Da-Wei and Li, Kai-Wen and Ning, Jingyi and Ye, Han-Jia and Zhang, Lijun and Zhan, De-Chuan},
  booktitle={Proceedings of the IEEE/CVF International Conference on Computer Vision},
  pages={3314--3325},
  year={2025}
}

@inproceedings{luo2025lada,
    title={{LADA}: Scalable Label-Specific {CLIP} Adapter for Continual Learning},
    author={Mao-Lin Luo and Zi-Hao Zhou and Tong Wei and Min-Ling Zhang},
    booktitle={Forty-second International Conference on Machine Learning},
    year={2025}
}

@article{gao2024clip,
  title={Clip-adapter: Better vision-language models with feature adapters},
  author={Gao, Peng and Geng, Shijie and Zhang, Renrui and Ma, Teli and Fang, Rongyao and Zhang, Yongfeng and Li, Hongsheng and Qiao, Yu},
  journal={International journal of computer vision},
  volume={132},
  number={2},
  pages={581--595},
  year={2024},
  publisher={Springer}
}

@article{zhou2022learning,
  title={Learning to prompt for vision-language models},
  author={Zhou, Kaiyang and Yang, Jingkang and Loy, Chen Change and Liu, Ziwei},
  journal={International journal of computer vision},
  volume={130},
  number={9},
  pages={2337--2348},
  year={2022},
  publisher={Springer}
}

@inproceedings{
wu2025sdlora,
title={{SD}-Lo{RA}: Scalable Decoupled Low-Rank Adaptation for Class Incremental Learning},
author={Yichen Wu and Hongming Piao and Long-Kai Huang and Renzhen Wang and Wanhua Li and Hanspeter Pfister and Deyu Meng and Kede Ma and Ying Wei},
booktitle={The Thirteenth International Conference on Learning Representations},
year={2025},
url={https://openreview.net/forum?id=5U1rlpX68A}
}

@inproceedings{ijcai2025p715,
  title     = {On the Discrimination and Consistency for Exemplar-Free Class Incremental Learning},
  author    = {Wang, Tianqi and Guo, Jingcai and Li, Depeng and Chen, Zhi},
  booktitle = {Proceedings of the Thirty-Fourth International Joint Conference on
               Artificial Intelligence, {IJCAI-25}},
  publisher = {International Joint Conferences on Artificial Intelligence Organization},
  editor    = {James Kwok},
  pages     = {6424--6432},
  year      = {2025},
  month     = {8},
  note      = {Main Track},
  doi       = {10.24963/ijcai.2025/715},
  url       = {https://doi.org/10.24963/ijcai.2025/715},
}

@article{zhang2024continual,
  title={Continual learning of image classes with language guidance from a vision-language model},
  author={Zhang, Wentao and Huang, Yujun and Zhang, Weizhuo and Zhang, Tong and Lao, Qicheng and Yu, Yue and Zheng, Wei-Shi and Wang, Ruixuan},
  journal={IEEE Transactions on Circuits and Systems for Video Technology},
  volume={34},
  number={12},
  pages={13152--13163},
  year={2024},
  publisher={IEEE}
}

@article{zhang2025visual,
  title={Visual class incremental learning with textual priors guidance based on an adapted vision-language model},
  author={Zhang, Wentao and Yu, Tong and Wang, Ruixuan and Xie, Jianhui and Trucco, Emanuele and Zheng, Wei-Shi and Yang, Xiaobo},
  journal={IEEE Transactions on Multimedia},
  year={2025},
  publisher={IEEE}
}

@article{liang2022mind,
  title={Mind the gap: Understanding the modality gap in multi-modal contrastive representation learning},
  author={Liang, Victor Weixin and Zhang, Yuhui and Kwon, Yongchan and Yeung, Serena and Zou, James Y},
  journal={Advances in Neural Information Processing Systems},
  volume={35},
  pages={17612--17625},
  year={2022}
}

@article{thengane2022clip,
  title={Clip model is an efficient continual learner},
  author={Thengane, Vishal and Khan, Salman and Hayat, Munawar and Khan, Fahad},
  journal={arXiv preprint arXiv:2210.03114},
  year={2022}
}

@inproceedings{zhang2023slca,
  title={Slca: Slow learner with classifier alignment for continual learning on a pre-trained model},
  author={Zhang, Gengwei and Wang, Liyuan and Kang, Guoliang and Chen, Ling and Wei, Yunchao},
  booktitle={Proceedings of the IEEE/CVF International Conference on Computer Vision},
  pages={19148--19158},
  year={2023}
}

@article{zhou2025learning,
  title={Learning without forgetting for vision-language models},
  author={Zhou, Da-Wei and Zhang, Yuanhan and Wang, Yan and Ning, Jingyi and Ye, Han-Jia and Zhan, De-Chuan and Liu, Ziwei},
  journal={IEEE Transactions on Pattern Analysis and Machine Intelligence},
  year={2025},
  publisher={IEEE}
}

@inproceedings{zheng2023preventing,
  title={Preventing zero-shot transfer degradation in continual learning of vision-language models},
  author={Zheng, Zangwei and Ma, Mingyuan and Wang, Kai and Qin, Ziheng and Yue, Xiangyu and You, Yang},
  booktitle={Proceedings of the IEEE/CVF international conference on computer vision},
  pages={19125--19136},
  year={2023}
}

@inproceedings{marczak2024magmax,
  title={Magmax: Leveraging model merging for seamless continual learning},
  author={Marczak, Daniel and Twardowski, Bart{\l}omiej and Trzci{\'n}ski, Tomasz and Cygert, Sebastian},
  booktitle={European Conference on Computer Vision},
  pages={379--395},
  year={2024},
  organization={Springer}
}

@inproceedings{eslami2025mitigate,
  title={Mitigate the gap: Improving cross-modal alignment in CLIP},
  author={Eslami, Sedigheh and de Melo, Gerard},
  booktitle={The Thirteenth International Conference on Learning Representations},
  year={2025}
}

@inproceedings{mistretta2025cross,
  title={Cross the Gap: Exposing the Intra-modal Misalignment in CLIP via Modality Inversion},
  author={Marco Mistretta and Alberto Baldrati and Lorenzo Agnolucci and Marco Bertini and Andrew D. Bagdanov},
  booktitle={The Thirteenth International Conference on Learning Representations},
  year={2025},
  url={https://openreview.net/forum?id=VVVfuIcmKR}
}

@inproceedings{schrodi2025two,
    title={Two Effects, One Trigger: On the Modality Gap, Object Bias, and Information Imbalance in Contrastive Vision-Language Models},
    author={Simon Schrodi and David T. Hoffmann and Max Argus and Volker Fischer and Thomas Brox},
    booktitle={The Thirteenth International Conference on Learning Representations},
    year={2025},
    url={https://openreview.net/forum?id=uAFHCZRmXk}
}

@inproceedings{cai2025misalignment,
  title     = {On the Value of Cross-Modal Misalignment in Multimodal Representation Learning},
  author    = {Cai, Yichao and Liu, Yuhang and Gao, Erdun and Jiang, Tianjiao and Zhang, Zhen and van den Hengel, Anton and Shi, Javen Qinfeng},
  booktitle = {Advances in Neural Information Processing Systems (NeurIPS)},
  year      = {2025}
}

@inproceedings{yamaguchi2025post,
  title={Post-pre-training for modality alignment in vision-language foundation models},
  author={Yamaguchi, Shin'ya and Feng, Dewei and Kanai, Sekitoshi and Adachi, Kazuki and Chijiwa, Daiki},
  booktitle={Proceedings of the Computer Vision and Pattern Recognition Conference},
  pages={4256--4266},
  year={2025}
}

@article{hu2022lora,
  title={Lora: Low-rank adaptation of large language models.},
  author={Hu, Edward J and Shen, Yelong and Wallis, Phillip and Allen-Zhu, Zeyuan and Li, Yuanzhi and Wang, Shean and Wang, Liang and Chen, Weizhu and others},
  journal={Iclr},
  volume={1},
  number={2},
  pages={3},
  year={2022}
}

@inproceedings{huang2024class,
  title={Class-incremental learning with clip: Adaptive representation adjustment and parameter fusion},
  author={Huang, Linlan and Cao, Xusheng and Lu, Haori and Liu, Xialei},
  booktitle={European Conference on Computer Vision},
  pages={214--231},
  year={2024},
  organization={Springer}
}

@article{zhou2025revisiting,
  title={Revisiting class-incremental learning with pre-trained models: Generalizability and adaptivity are all you need},
  author={Zhou, Da-Wei and Cai, Zi-Wen and Ye, Han-Jia and Zhan, De-Chuan and Liu, Ziwei},
  journal={International Journal of Computer Vision},
  volume={133},
  number={3},
  pages={1012--1032},
  year={2025},
  publisher={Springer}
}

@article{hendrycks2016baseline,
  title={A baseline for detecting misclassified and out-of-distribution examples in neural networks},
  author={Hendrycks, Dan and Gimpel, Kevin},
  journal={arXiv preprint arXiv:1610.02136},
  year={2016}
}

@inproceedings{bossard2014food,
  title={Food-101--mining discriminative components with random forests},
  author={Bossard, Lukas and Guillaumin, Matthieu and Van Gool, Luc},
  booktitle={European conference on computer vision},
  pages={446--461},
  year={2014},
  organization={Springer}
}

@inproceedings{parkhi2012cats,
  title={Cats and dogs},
  author={Parkhi, Omkar M and Vedaldi, Andrea and Zisserman, Andrew and Jawahar, CV},
  booktitle={2012 IEEE conference on computer vision and pattern recognition},
  pages={3498--3505},
  year={2012},
  organization={IEEE}
}

@inproceedings{liu2025c,
  title={C-CLIP: Multimodal continual learning for vision-language model},
  author={Liu, Wenzhuo and Zhu, Fei and Wei, Longhui and Tian, Qi},
  booktitle={The Thirteenth International Conference on Learning Representations},
  year={2025}
}

@ARTICLE{2025arXiv250910535L,
       author = {{Li}, Miaoge and {Chen}, Yang and {Rao}, Zhijie and {Jiang}, Can and {Guo}, Jingcai},
        title = "{Semantic-guided LoRA Parameters Generation}",
      journal = {arXiv e-prints},
         year = 2025,
        month = sep,
          eid = {arXiv:2509.10535},
        pages = {arXiv:2509.10535},
          doi = {10.48550/arXiv.2509.10535},
archivePrefix = {arXiv},
       eprint = {2509.10535},
 primaryClass = {stat.ML},
       adsurl = {https://ui.adsabs.harvard.edu/abs/2025arXiv250910535L}
}

\newpage
\appendix

\section{Theoretical Proofs and Clarification}\label{proof}

\paragraph{Feasible set in the text subspace.}
For task $t$, let the text feature matrix be $\mathbf{T}^t = \mathbf{U}_t \mathbf{\Sigma}_t \mathbf{V}_t^\top$, and let
$\mathbf{P}_t = \mathbf{U}_t \mathbf{U}_t^\top$ be the orthogonal projector onto the text subspace
$\mathrm{span}(\mathbf{U}_t)$. Define the feasible set of linear classifiers constrained to the text subspace as:
\[
\mathcal{W}_t^{\mathrm{text}}
=
\left\{
W \in \mathbb{R}^{d \times |\mathcal{C}^t|} \;:\; \operatorname{col}(W)\subseteq \operatorname{span}(\mathbf{U}_t)
\right\}.
\]

\subsection{Proof of Proposition~\ref{proposition_1}.}\label{proof_pro1}
\paragraph{Proof.} Recall Proposition~\ref{proposition_1}: let $W_t^\star$ denote an ideal linear classifier in the full visual feature space,
used only for analysis. If the classifier is constrained to lie in the text subspace, then its best approximation is:
\[
W_{t,\mathrm{text}}^\star = \mathbf{P}_t W_t^\star,
\]
and the corresponding approximation error is
\[
\mathcal{E}_{\mathrm{text}}^t = \|(\mathbf{I}-\mathbf{P}_t)W_t^\star\|_F^2.
\]

We consider the constrained approximation problem
\[
\min_{W \in \mathcal{W}_t^{\mathrm{text}}} \|W - W_t^\star\|_F^2.
\]
Since $\mathbf{P}_t$ is the orthogonal projector onto $\operatorname{span}(\mathbf{U}_t)$, we decompose $W_t^\star$ into its projection onto the text subspace and its orthogonal residual:
\[
W_t^\star = \mathbf{P}_t W_t^\star + (\mathbf{I}-\mathbf{P}_t)W_t^\star.
\]
Therefore,
\[
W - W_t^\star
= (W - \mathbf{P}_t W_t^\star) - (\mathbf{I}-\mathbf{P}_t)W_t^\star.
\]
By orthogonality, the Frobenius norm admits the Pythagorean decomposition:
\[
\|W - W_t^\star\|_F^2
=
\|W - \mathbf{P}_t W_t^\star\|_F^2
+
\|(\mathbf{I}-\mathbf{P}_t)W_t^\star\|_F^2.
\]
The second term is independent of $W$, and the first term is minimized if and only if:
\[
W = \mathbf{P}_t W_t^\star.
\]
Hence the best approximation in the text subspace is:
\[
W_{t,\mathrm{text}}^\star = \mathbf{P}_t W_t^\star,
\]
and the optimal approximation error is:
\[
\mathcal{E}_{\mathrm{text}}^t
=
\min_{W \in \mathcal{W}_t^{\mathrm{text}}} \|W - W_t^\star\|_F^2
=
\|(\mathbf{I}-\mathbf{P}_t)W_t^\star\|_F^2.
\]
This completes the proof.

\subsection{Proof of Lemma~\ref{lemma1}.}
\paragraph{Proof.} Recall that for task $t$, the text feature matrix admits the SVD
$\mathbf{T}^t = \mathbf{U}_t \mathbf{\Sigma}_t \mathbf{V}_t^\top$, and the orthogonal projector onto the text subspace is
$\mathbf{P}_t = \mathbf{U}_t \mathbf{U}_t^\top$. Let $r_t = \operatorname{rank}(\mathbf{T}^t)$ and
$\rho_t = \operatorname{rank}(W_t^\star)$. Denote the singular values of $W_t^\star$ by:
\[
\sigma_1(W_t^\star)\ge \sigma_2(W_t^\star)\ge \cdots \ge \sigma_{\rho_t}(W_t^\star)>0.
\]

By Proposition~\ref{proposition_1}, the text-subspace approximation error is:
\[
\mathcal{E}_{\mathrm{text}}^t
=
\|(\mathbf{I}-\mathbf{P}_t)W_t^\star\|_F^2 .
\]
Since $\mathbf{P}_t$ is an orthogonal projector, we have:
\[
\|(\mathbf{I}-\mathbf{P}_t)W_t^\star\|_F^2
=
\|W_t^\star\|_F^2-\|\mathbf{P}_tW_t^\star\|_F^2.
\]
Next, define the positive semidefinite matrix:
\[
M_t := W_t^\star W_t^{\star\top}\succeq 0.
\]
Then,
\begin{align}
\|P_tW_t^\star\|_F^2
&=
\operatorname{tr}\!\left((\mathbf{P}_tW_t^\star)^\top(\mathbf{P}_tW_t^\star)\right)\notag \\
&=
\operatorname{tr}\!\left(W_t^{\star\top}\mathbf{P}_tW_t^\star\right)\notag \\
&=
\operatorname{tr}\!\left(\mathbf{P}_tW_t^\star W_t^{\star\top}\right)\notag \\
&=
\operatorname{tr}(\mathbf{P}_tM_t)\notag.
\end{align}
Note that $M_t$ has eigenvalues:
\[
\lambda_j(M_t)=\sigma_j^2(W_t^\star), \quad j=1,\dots,\rho_t,
\]
with the remaining eigenvalues being zero. Since $P_t$ is a rank-$r_t$ orthogonal projector,
by the equivalent trace maximization result for symmetric positive semidefinite matrices,
\begin{equation}
\operatorname{tr}(\mathbf{P}_tM_t)
\le
\sum_{j=1}^{r_t}\lambda_j(M_t)
=
\sum_{j=1}^{r_t}\sigma_j^2(W_t^\star).
\end{equation}
Therefore,
\begin{align}
\mathcal{E}_{\mathrm{text}}^t
&=
\|W_t^\star\|_F^2-\operatorname{tr}(\mathbf{P}_tM_t)\notag \\
&\ge
\sum_{j=1}^{\rho_t}\sigma_j^2(W_t^\star)
-
\sum_{j=1}^{r_t}\sigma_j^2(W_t^\star)\notag \\
&=
\sum_{j=r_t+1}^{\rho_t}\sigma_j^2(W_t^\star)\notag.
\end{align}
This proves the desired lower bound. Moreover, equality holds if and only if
\[
\operatorname{tr}(\mathbf{P}_tM_t)=\sum_{j=1}^{r_t}\lambda_j(M_t),
\]
which is equivalent to $\mathbf{P}_t$ projecting onto the eigenspace associated with the top $r_t$
eigenvalues of $M_t$. Since the eigenvectors of $M_t=W_t^\star W_t^{\star\top}$ are exactly
the left singular vectors of $W_t^\star$, equality holds if and only if the text subspace
$\operatorname{span}(\mathbf{P}_t)$ covers the leading $r_t$ left singular directions of $W_t^\star$.
This completes the proof.

\paragraph{Feasible set in the direct-sum subspace.}
Let $S_R^t \subseteq \operatorname{span}(\mathbf{P}_t^\perp)$ be a compensation subspace for task $t$,
and let $\mathbf{P}_{R,t}$ denote the orthogonal projector onto $S_R^t$.
Define the feasible set of classifiers constrained to the direct-sum subspace
$\operatorname{span}(\mathbf{P}_t)\oplus S_R^t$ as:
\[
\mathcal{W}_t^\oplus
=
\left\{
W \in \mathbb{R}^{d \times |\mathcal{C}^t|}\;:\;
\operatorname{col}(W)\subseteq \operatorname{span}(\mathbf{P}_t)\oplus S_R^t
\right\}.
\]

\subsection{Proof of Proposition~\ref{proposition_2}.}
\paragraph{Proof.} Since $S_R^t \subseteq \operatorname{span}(\mathbf{P}_t^\perp)$, the two subspaces
$\operatorname{span}(\mathbf{P}_t)$ and $S_R^t$ are orthogonal. Therefore,
\[
\mathbf{P}_t\mathbf{P}_{R,t}=\mathbf{P}_{R,t}\mathbf{P}_t=0,
\]
and $\mathbf{P}_t+\mathbf{P}_{R,t}$ is exactly the orthogonal projector onto the direct-sum subspace
$\operatorname{span}(\mathbf{P}_t)\oplus S_R^t$.

Consider the constrained approximation problem:
\[
\min_{W\in\mathcal{W}_t^\oplus}\|W-W_t^\star\|_F^2.
\]
For any feasible $W\in\mathcal{W}_t^\oplus$, we have:
\[
W=(\mathbf{P}_t+\mathbf{P}_{R,t})W.
\]
Decompose the ideal classifier $W_t^\star$ into its projection onto the direct-sum subspace
and its orthogonal residual:
\[
W_t^\star
=
(\mathbf{P}_t+\mathbf{P}_{R,t})W_t^\star
+
(\mathbf{I}-\mathbf{P}_t-\mathbf{P}_{R,t})W_t^\star.
\]
Hence,
\[
W-W_t^\star
=
\big(W-(\mathbf{P}_t+\mathbf{P}_{R,t})W_t^\star\big)
-
(\mathbf{I}-\mathbf{P}_t-\mathbf{P}_{R,t})W_t^\star.
\]
By the Pythagorean theorem for the Frobenius norm,
\[
\|W-W_t^\star\|_F^2
=
\|W-(\mathbf{P}_t+\mathbf{P}_{R,t})W_t^\star\|_F^2
+
\|(\mathbf{I}-\mathbf{P}_t-\mathbf{P}_{R,t})W_t^\star\|_F^2.
\]
The second term is independent of $W$, and the first term is minimized if and only if:
\[
W=(\mathbf{P}_t+\mathbf{P}_{R,t})W_t^\star.
\]
Therefore, the best approximation of $W_t^\star$ in the direct-sum subspace is:
\[
W_{t,\oplus}^\star=(\mathbf{P}_t+\mathbf{P}_{R,t})W_t^\star,
\]
with approximation error:
\[
\mathcal{E}_\oplus^t
=
\min_{W\in\mathcal{W}_t^\oplus}\|W-W_t^\star\|_F^2
=
\|(\mathbf{I}-\mathbf{P}_t-\mathbf{P}_{R,t})W_t^\star\|_F^2.
\]
This completes the proof.

\subsection{Proof of Corollary~\ref{corollary}.}
\paragraph{Proof.} By Proposition~\ref{proposition_1} and Proposition~\ref{proposition_2}, we have
\[
\mathcal{E}_{\mathrm{text}}^t=\|(\mathbf{I}-\mathbf{P}_t)W_t^\star\|_F^2,
\qquad
\mathcal{E}_\oplus^t=\|(\mathbf{I}-\mathbf{P}_t-\mathbf{P}_{R,t})W_t^\star\|_F^2.
\]
Since $S_R^t \subseteq \operatorname{span}(\mathbf{P}_t^\perp)$, the projector $\mathbf{P}_{R,t}$
acts within the orthogonal complement of $\operatorname{span}(\mathbf{P}_t)$. Thus,
\[
(\mathbf{I}-\mathbf{P}_t)W_t^\star
=
\mathbf{P}_{R,t}(\mathbf{I}-\mathbf{P}_t)W_t^\star
+
(\mathbf{I}-\mathbf{P}_t-\mathbf{P}_{R,t})W_t^\star,
\]
where the first term lies in $S_R^t$ and the second term lies in the orthogonal complement
of $\operatorname{span}(\mathbf{P}_t)\oplus S_R^t$. These two terms are column-wise orthogonal, and hence,
\[
\|(\mathbf{I}-\mathbf{P}_t)W_t^\star\|_F^2
=
\|\mathbf{P}_{R,t}(\mathbf{I}-\mathbf{P}_t)W_t^\star\|_F^2
+
\|(\mathbf{I}-\mathbf{P}_t-\mathbf{P}_{R,t})W_t^\star\|_F^2.
\]
Therefore,
\[
\mathcal{E}_{\mathrm{text}}^t-\mathcal{E}_\oplus^t
=
\|\mathbf{P}_{R,t}(\mathbf{I}-\mathbf{P}_t)W_t^\star\|_F^2
\ge 0,
\]
which immediately implies:
\[
\mathcal{E}_\oplus^t\le \mathcal{E}_{\mathrm{text}}^t.
\]
This completes the proof.

The above result can be further characterized in terms of the singular-value tail energy of $W_t^\star$.

\subsection{Singular-value form of the direct-sum approximation error.}
Let:
\[
r_t=\operatorname{rank}(\mathbf{T}^t),\qquad
k_t=\operatorname{rank}(\mathbf{P}_{R,t})=\dim(S_R^t),\qquad
m_t=r_t+k_t.
\]
Then $\mathbf{P}_t+\mathbf{P}_{R,t}$ is a rank-$m_t$ orthogonal projector. Let
$\rho_t=\operatorname{rank}(W_t^\star)$, and denote the singular values of $W_t^\star$ by:
\[
\sigma_1(W_t^\star)\ge \sigma_2(W_t^\star)\ge \cdots \ge \sigma_{\rho_t}(W_t^\star)>0.
\]
Then,
\[
\mathcal{E}_\oplus^t
=
\|(\mathbf{I}-\mathbf{P}_t-\mathbf{P}_{R,t})W_t^\star\|_F^2
\ge
\sum_{j=m_t+1}^{\rho_t}\sigma_j^2(W_t^\star).
\]

\paragraph{Proof.}
Define $M_t=W_t^\star W_t^{\star\top}\succeq 0$. Since $\mathbf{P}_t+\mathbf{P}_{R,t}$ is a rank-$m_t$
orthogonal projector, we have:
\begin{align}
\mathcal{E}_\oplus^t
&=
\|W_t^\star\|_F^2-\|(\mathbf{P}_t+\mathbf{P}_{R,t})W_t^\star\|_F^2 \notag\\
&=
\|W_t^\star\|_F^2-\operatorname{tr}\big((\mathbf{P}_t+\mathbf{P}_{R,t})M_t\big).\notag
\end{align}
By the equivalent trace maximization result,
\[
\operatorname{tr}\big((\mathbf{P}_t+\mathbf{P}_{R,t})M_t\big)
\le
\sum_{j=1}^{m_t}\lambda_j(M_t)
=
\sum_{j=1}^{m_t}\sigma_j^2(W_t^\star).
\]
Substituting this into the above identity yields:
\[
\mathcal{E}_\oplus^t
\ge
\sum_{j=1}^{\rho_t}\sigma_j^2(W_t^\star)
-
\sum_{j=1}^{m_t}\sigma_j^2(W_t^\star)
=
\sum_{j=m_t+1}^{\rho_t}\sigma_j^2(W_t^\star).
\]
Equality holds if and only if the direct-sum subspace
$\operatorname{span}(\mathbf{P}_t)\oplus S_R^t$ covers the leading $m_t$ left singular directions
of $W_t^\star$.

\subsection{Singular-value upper bound on the error reduction.}
Under the same notation, the improvement brought by orthogonal compensation satisfies
\[
\mathcal{E}_{\mathrm{text}}^t-\mathcal{E}_\oplus^t
\le
\sum_{j=r_t+1}^{m_t}\sigma_j^2(W_t^\star).
\]

\paragraph{Proof.}
\begin{align}
\mathcal{E}_{\mathrm{text}}^t-\mathcal{E}_\oplus^t
&\le
\sum_{j=r_t+1}^{\rho_t}\sigma_j^2(W_t^\star)
-
\sum_{j=m_t+1}^{\rho_t}\sigma_j^2(W_t^\star) \notag\\
&=
\sum_{j=r_t+1}^{m_t}\sigma_j^2(W_t^\star)\notag.
\end{align}
This proves the claim.

\subsection{Theory-to-practice clarification.}
The theoretical results are intended as a geometric justification of the proposed design, rather than an exact description of the optimization procedure. In particular, the classifier $W_t^\star$ is introduced only as an ideal linear classifier in the full visual feature space for analysis, and is not explicitly learned in practice. Proposition~\ref{proposition_1} and Lemma~\ref{lemma1} show that, if classification is restricted to the text subspace, the uncovered discriminative energy is quantified by the projection residual and its singular-value tail. Proposition~\ref{proposition_2} and Corollary~\ref{corollary} further show that introducing an additional subspace in the orthogonal complement can reduce this approximation error.

In practice, for task $t$, the text subspace is constructed from the current task text feature matrix $\mathbf{T}^t$. Specifically, we compute the SVD of $\mathbf{T}^t$ and use its left singular vectors to form the projector $\mathbf{P}_t = \mathbf{U}_t\mathbf{U}_t^\top$. The compensation head is then parameterized as:
\[
\widehat{W}_{\mathrm{comp}}^t = \mathbf{P}_t^\perp W_{\mathrm{comp}}^t,
\]
so that its column space is constrained to lie in the orthogonal complement of the text subspace. In this way, the learnable compensation head serves as a practical parameterization of the residual modeling discussed in the theory.

We emphasize that the theory does not claim that the learned compensation head exactly recovers the optimal residual subspace or achieves the singular-value bound in practice. Instead, it explains why modeling an additional classifier in the orthogonal complement is well motivated whenever the text subspace does not fully cover the discriminative structure of the visual space.

\section{Algorithms Details}\label{Algorithms}

\subsection{Training algorithm of GR4CIL.}

\paragraph{Training procedure.}
For each task, GR4CIL is trained in two stages. We first learn the current task-specific visual LoRA together with the shared text LoRA using the base objective in Eq.~\ref{eq3}, while freezing previously learned visual LoRAs. After that, we freeze the learned visual and text branches, and train the orthogonal compensation head using Eq.~\ref{eq7}.

\begin{algorithm}[h]
\caption{Training Procedure of GR4CIL}
\label{alg:training_gr4cil}
\begin{algorithmic}[1]
\REQUIRE Task sequence $\{\mathcal{D}^1,\dots,\mathcal{D}^T\}$; pretrained CLIP $(f_v,f_t)$; shared text LoRA $\phi_{\mathrm{text}}$; task-specific visual LoRA bank $\{\phi_{\mathrm{vis}}^t\}_{t=1}^T$
\ENSURE Learned $\phi_{\mathrm{text}}$; task-specific $\{\phi_{\mathrm{vis}}^t,\widehat{W}_{\mathrm{comp}}^t,\mathbf{p}_c^t\}_{t=1}^T$; cached text anchors $\{{\mathbf{z}}^c\}$

\STATE Initialize shared text LoRA $\phi_{\mathrm{text}}$ and empty memory for text anchors
\FOR{$t=1$ to $T$}
    \STATE Initialize current visual LoRA $\phi_{\mathrm{vis}}^t$
    \STATE Freeze all previous visual LoRAs $\{\phi_{\mathrm{vis}}^\tau\}_{\tau=1}^{t-1}$

    \STATE \textbf{Stage 1: Incremental knowledge learning}
    \FOR{$e=1$ to $Epoch_{\mathrm{base}}$}
        \FOR{each mini-batch $(\mathbf{x}_i,y_i)$ in $\mathcal{D}^t$}
        \STATE Compute current-task visual features $\mathbf{v}_i^t=f_v(\mathbf{x}_i;\phi_{\mathrm{vis}}^t)$
        \STATE Compute text features $\{\mathbf{t}_c^t\}_{c\in\mathcal{C}^{1:t}}$ with shared text LoRA $\phi_{\mathrm{text}}$
        \STATE Compute $\mathcal{L}_{\mathrm{anc}}$ using cached anchors of previous classes
        \STATE Compute $\mathcal{L}_{\mathrm{sep}}$ over current-task classes
        \STATE Compute $\mathcal{L}_{\mathrm{base}}=\mathcal{L}_{\mathrm{clip}}+\lambda_{\mathrm{anc}}\mathcal{L}_{\mathrm{anc}}+\lambda_{\mathrm{sep}}\mathcal{L}_{\mathrm{sep}}$
        \STATE Update $\phi_{\mathrm{vis}}^t$ and $\phi_{\mathrm{text}}$
        \ENDFOR
    \ENDFOR

    \STATE Cache text anchors ${\mathbf{z}}^c$ for newly learned classes $c\in\mathcal{C}^t$
    \STATE Compute visual class prototypes $\{\mathbf{p}_c^t\}_{c\in\mathcal{C}^t}$ from current-task features

    \STATE \textbf{Stage 2: Orthogonal compensation learning}
    \STATE Freeze $\phi_{\mathrm{vis}}^t$ and $\phi_{\mathrm{text}}$
    \STATE Build text feature matrix $\mathbf{T}^t=[\mathbf{t}_{c_1}^t,\dots,\mathbf{t}_{c_{|\mathcal{C}^t|}}^t]$
    \STATE Compute $\mathbf{P}_t=\mathbf{U}_t\mathbf{U}_t^\top$ from the SVD of $\mathbf{T}^t$, and set $\mathbf{P}_t^\perp=\mathbf{I}-\mathbf{P}_t$
    \STATE Initialize $W_{\mathrm{comp}}^t$ with current-task visual prototypes $\{\mathbf{p}_c^t\}_{c\in\mathcal{C}^t}$

    \FOR{$e=1$ to $Epoch_{\mathrm{comp}}$}
        \FOR{each mini-batch $(\mathbf{x}_i,y_i)$ in $\mathcal{D}^t$}
        \STATE Compute $\mathbf{v}_i^t=f_v(\mathbf{x}_i;\phi_{\mathrm{vis}}^t)$
        \STATE Compute $\widehat{W}_{\mathrm{comp}}^t=P_t^\perp W_{\mathrm{comp}}^t$
        \STATE Compute compensation logits $g^t(\mathbf{x}_i)=\mathbf{v}_i^{t\top}\widehat{W}_{\mathrm{comp}}^t$
        \STATE Compute $\mathcal{L}_{\mathrm{comp}}$
        \STATE Update $W_{\mathrm{comp}}^t$
        \ENDFOR
    \ENDFOR

    \STATE Save $\phi_{\mathrm{vis}}^t$, $\widehat{W}_{\mathrm{comp}}^t$, and $\{\mathbf{p}_c^t\}_{c\in\mathcal{C}^t}$
\ENDFOR
\end{algorithmic}
\end{algorithm}

\subsection{Unified inference for CIL.}

\paragraph{Unified inference.}
Under the standard CIL setting, GR4CIL performs closed-set prediction over all seen classes via unified score competition across task-specific branches. For an input sample, each learned task-specific visual branch produces a task-conditioned visual feature, which is then combined with the text classifier, orthogonal compensation head, and prototype term to form the final class score. The prediction is obtained by taking the maximum score over all seen classes.

\begin{algorithm}[h]
\caption{Unified Inference of GR4CIL}
\label{alg:inference_gr4cil}
\begin{algorithmic}[1]
\REQUIRE Test sample $\mathbf{x}$; learned shared text branch; task-specific visual LoRAs $\{\phi_{\mathrm{vis}}^t\}_{t=1}^T$; compensation heads $\{\widehat{W}_{\mathrm{comp}}^t\}_{t=1}^T$; class prototypes $\{\mathbf{p}_c^t\}$; seen class sets $\{\mathcal{C}^t\}_{t=1}^T$
\ENSURE Predicted label $\widehat{y}$

\FOR{$t=1$ to $T$}
    \STATE Compute task-conditioned visual feature $\mathbf{v}^t=f_v(\mathbf{x};\phi_{\mathrm{vis}}^t)$
    \FOR{each class $c\in\mathcal{C}^t$}
        \STATE Compute text score $s_c(\mathbf{x})=\langle \mathbf{v}^t,\mathbf{t}_c\rangle$
    \ENDFOR
    \STATE Compute compensation logits $\mathbf{g}^t(\mathbf{x})=\mathbf{v}^{t\top}\widehat{W}_{\mathrm{comp}}^t$
    \FOR{each class $c\in\mathcal{C}^t$}
        \STATE Compute compensated score $\widehat{s}_c(\mathbf{x})=s_c(\mathbf{x})+\beta\, g_c^t(\mathbf{x})$
        \STATE Compute final score $q_c(\mathbf{x})=\widehat{s}_c(\mathbf{x})+\gamma\,\langle \mathbf{v}^t,\mathbf{p}_c^t\rangle$
    \ENDFOR
\ENDFOR

\STATE $\widehat{y}=\arg\max_{c\in\mathcal{C}^{1:T}} q_c(\mathbf{x})$
\RETURN $\widehat{y}$
\end{algorithmic}
\end{algorithm}

\subsection{Algorithm interfaces for more open scenarios.}

It should be emphasized that the core formulation of GR4CIL still corresponds to the standard CIL setting. The training and unified inference procedures described in Appendix~\ref{Algorithms} already constitute the main body of the proposed method. Beyond this, GR4CIL naturally leaves a confidence-based extension interface: when none of the learned task branches can provide sufficiently reliable responses to an input, the model may further extend inference toward a more general prediction mode, rather than being restricted to closed-set decisions within the learned label space.

A prior work, MoE-Adapters~\citep{yu2024boosting}, introduced the Distribution Discriminative Auto-Selector (DDAS), whose core idea is likewise to first determine whether a sample can be sufficiently explained by the currently learned tasks based on distribution-aware confidence. When the existing task knowledge is insufficient, the sample is then routed to the frozen CLIP for zero-shot prediction. Similar to that work, we do not view this component as a complete solution to open-world inference, but rather as a practical interface beyond the standard CIL pipeline. The key difference is that DDAS relies on additionally trained task-specific autoencoders and a reference autoencoder for routing, whereas GR4CIL directly reuses the task-aware scores and prototype-based confidence already produced during standard CIL inference, so that this interface remains unified with the main routing mechanism. In addition, we further introduce a knowledge-fusion strategy, which enhances zero-shot generalization to a certain extent.

More specifically, GR4CIL further interprets the task-aware scores and prototype-based confidence as a task-relative OOD signal. For a given task branch, samples from other tasks can themselves be regarded as relatively OOD. Based on this view, when all task branches fail to produce sufficiently confident responses, GR4CIL can optionally trigger a generalized prediction branch that performs knowledge-fusion prediction over a given candidate label set. If this extension branch is not used, the original CLIP can still be naturally retained as a fallback predictor.

Therefore, from the perspective of positioning, we emphasize that GR4CIL leaves a practical interface beyond standard CIL, rather than directly claiming to have fully solved open-world inference. The OOD detection and zero-shot generalization experiments in the main text are primarily intended to demonstrate the potential feasibility of this interface: the former shows that the model can indeed provide a relatively reliable trigger signal, while the latter indicates that, once the interface is activated, the model still retains a certain level of generalized prediction capability. A more detailed limitation is provided in Appendix~\ref{Limitation}.

\section{Implementation and Evaluation Details}\label{Implementation}

\subsection{Implementation details.}

Unless otherwise specified, we largely follow the training protocol of AdapterVLM~\citep{zhang2025visual} in our implementation details, including the same CLIP ViT-B/16 backbone, LoRA architecture, visual-side data augmentation, text-side prompt design, and optimizer configuration for LoRA training.

All input images are resized to $224\times224$. During training, the data augmentation consists of random horizontal flipping, and random rotation with an angle range of $[0^\circ,10^\circ]$. During evaluation, images are resized to 
$224\times224$ without stochastic augmentation. 


For incremental knowledge learning, the LoRA rank is set to 24 for both the visual and textual branches. We use AdamW as the optimizer with a learning rate of 0.005 and a cosine annealing schedule. The balancing coefficients $\lambda_{\mathrm{anc}}$ and $\lambda_{\mathrm{sep}}$ are both set to 1, and the separation threshold $\tau$ is set to 0.7. For CIFAR100 and ImageNet-R, each task is trained for 70 epochs, while for ImageNet100 and ImageNet-1K each task is trained for 10 epochs. The batch size is set to 64 for all experiments.

For orthogonal compensation learning, after completing the LoRA training we freeze the current visual branch and the shared text branch, and train the compensation head separately. The compensation head is optimized by Adam with a learning rate of 0.0005. For CIFAR100 and ImageNet-R, the compensation head is trained for 3 epochs per task, while for ImageNet100 and ImageNet-1K it is trained for 5 epochs per task. During inference, the coefficients $\beta$ and $\gamma$ are both set to 0.2. Task prototypes are constructed from the mean of normalized visual features of the current task, and both prototypes and features are $L_2$-normalized before use. All experiments are conducted on one NVIDIA RTX 4090 GPU.

\subsection{Baseline sources and reproduction details.}

For fair comparison, we mainly refer to the reported results of MG-CLIP~\citep{huang2025mind} and AdapterVLM~\citep{zhang2025visual}, while reproducing the missing or unmatched settings using their released codebases. In particular, AdapterVLM originally reports results under the mean class recall (MCR) metric, which are not fully consistent with our setting. At the same time, it has not been tested on ImageNet100 and ImageNet-1K. Therefore, we reproduce AdapterVLM using its source code and evaluate it with the same metrics as in the main text. In all comparisons, we keep the CLIP ViT-B/16 backbone unchanged to reduce discrepancies caused by differences in backbone or evaluation protocol.

\subsection{Details of subspace-distance metric.}

To analyze the geometric relationship between the text classifier, the compensation classifier, and the principal discriminative directions of image features, we further compute several directional subspace-distance metrics. Specifically, we first extract orthonormal bases from the image features, the text classifier, and the compensation classifier. For the image features, we first apply $L_2$ normalization to each sample feature, and then perform SVD to obtain a principal subspace that preserves 95\% of the cumulative energy, denoted by $B_i$. For the text classifier matrix and the compensation matrix, we treat their row vectors as discriminative directions and extract orthonormal bases via SVD, denoted by $B_t$ and $B_c$, respectively. The joint subspace $B_{t\cup c}$ is obtained by concatenating $B_t$ and $B_c$ and then re-orthogonalizing the resulting basis.

Based on these bases, we define the directional distance as:
\[
d(B_{\mathrm{src}}, B_{\mathrm{tgt}})
=
\frac{1}{r_{\mathrm{src}}}
\sum_{j=1}^{r_{\mathrm{src}}}
\left\|
b_j - P_{\mathrm{tgt}} b_j
\right\|_2,
\qquad
P_{\mathrm{tgt}} = B_{\mathrm{tgt}} B_{\mathrm{tgt}}^\top,
\]
where \(B_{\mathrm{src}}=[b_1,\dots,b_{r_{\mathrm{src}}}]\) is an orthonormal basis of the source subspace and \(P_{\mathrm{tgt}}\) is the orthogonal projector onto the target subspace. This metric measures how much the directions in the source subspace cannot be explained by the target subspace. Therefore, a smaller value indicates that the target subspace better covers the principal directions of the source subspace. Note that this metric is directional and is generally not symmetric, i.e., \(d(B_1,B_2)\neq d(B_2,B_1)\).

In this paper, we mainly report \(d(B_i,B_t)\) (I-T), \(d(B_i,B_c)\) (I-C), and \(d(B_i,B_{t\cup c})\) (I-TC). Here, \(d(B_i,B_t)\) characterizes the mismatch between the image discriminative subspace and the text classifier subspace, while \(d(B_i,B_{t\cup c})\) measures how well the joint text-compensation space covers the image discriminative directions. A smaller \(d(B_i,B_{t\cup c})\) than \(d(B_i,B_t)\) indicates that the proposed compensation mechanism effectively reduces the geometric discrepancy between the image space and the text classifier space. See Table~\ref{tab:analysis_tables} (right) in the main text for the results.

\subsection{Details of computational cost metric.}

The computational cost reported in the main text mainly includes two metrics: the number of trainable parameters and the GPU memory usage. For the parameter metric, we count the total number of trainable parameters throughout the whole incremental process. Specifically, for our method, the shared text LoRA is counted only once, while the task-specific visual LoRAs and the compensation heads are accumulated across tasks. Additional storage such as prototypes and cached text anchors is also included in the parameter count. For the memory metric, we report the maximum peak GPU memory observed over both training and inference throughout the entire incremental process. All methods are measured under the same batch size to reduce discrepancies caused by implementation settings.

\section{Additional Routing Analysis}\label{routing}

This section provides a more direct evaluation of routing behavior under the standard CIL setting. While the main text analyzes routing through score margins and compensation effects, here we explicitly define routing accuracy and examine how it evolves throughout the incremental process.

\subsection{Definition of routing accuracy.}
In our framework, routing is not performed by an additional task predictor, but is instead implicitly realized through unified class-level score competition across all learned task-specific branches. For a test sample $\mathbf{x}_i$, the model first produces the final prediction over all learned classes:
\[
\widehat{y}_i=\arg\max_{c\in\mathcal{C}^{1:T}} q_c(\mathbf{x}_i),
\]
where $q_c(\mathbf{x}_i)$ denotes the final class score in the unified inference rule of the main text. Since each class $c$ belongs to one and only one incremental task, the predicted class $\widehat{y}_i$ naturally determines a predicted task, denoted by $t(\widehat{y}_i)$. Similarly, the GT label $y_i$ belongs to the task $t(y_i)$.

Based on this, we define the routing accuracy as whether the task implied by the final prediction matches the task of the GT class:
\[
\mathrm{RA}
=
\frac{1}{N}\sum_{i=1}^{N}\mathbf{1}\!\left[t(\widehat{y}_i)=t(y_i)\right],
\]
where $N$ is the number of test samples and $\mathbf{1}(\cdot)$ is the indicator function.

This definition does not require an additional task-level classifier, since under the standard CIL setting the predicted task is uniquely determined by the predicted class. Compared with only analyzing the score margin between the GT task and competing tasks, routing accuracy more directly measures whether a sample is assigned to the correct task branch. Therefore, it serves as a direct complement to the margin analysis in the main text and allows us to verify whether the proposed compensation mechanism truly improves task-level discrimination under unified inference.

\subsection{Per-stage routing curves.}

We analyze the stage-wise evolution of routing accuracy on CIFAR100, ImageNet-R, and ImageNet100 under the 10-step incremental setting. Specifically, we compare three variants: (1) using only the text classifier, (2) augmenting the text classifier with the prototype-based OOD term, and (3) the full model with both the prototype-based OOD term and the compensation module.

\begin{figure}[t]
  \centering
  \begin{subfigure}[t]{0.98\linewidth}
    \includegraphics[width=\linewidth]{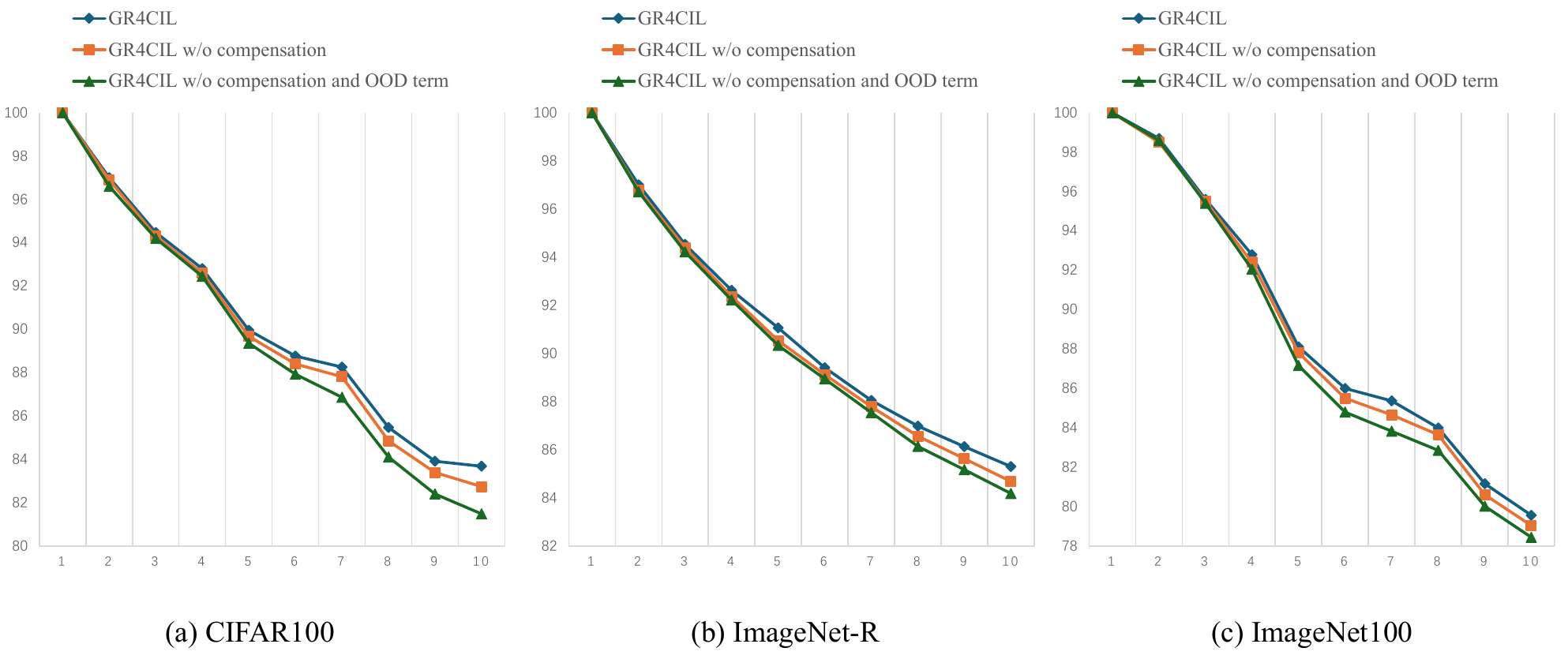}
  \end{subfigure}
  \caption{Per-stage routing accuracy on CIFAR100, ImageNet-R, and ImageNet100 under the 10-step setting. We compare three variants: text classifier only, text classifier with the OOD term, and the full model with both the OOD term and the compensation module. The results show that the OOD term already improves routing, while the compensation module further enhances task-level score separability and leads to more accurate routing across stages.}
  \label{fig7}
\end{figure}

Fig.~\ref{fig7} show that the original text classifier consistently yields the lowest routing accuracy across all three datasets, suggesting that relying only on the text classifier is insufficient to provide clear score separability among competing task branches. After introducing the prototype-based OOD term, the routing accuracy is improved at most stages, indicating that the task-internal confidence induced by prototype similarity indeed helps distinguish in-distribution samples of the current task from task-relative OOD samples coming from other learned tasks.

On top of this, the full model with the compensation module further improves routing accuracy, and such improvements remain relatively stable across incremental stages. This suggests that orthogonal compensation not only improves intra-task discrimination, but also strengthens the score boundary between the GT task and competing tasks, thereby promoting clearer score separation across different task branches. Overall, these stage-wise results are consistent with the margin analysis in the main text and further validate the effectiveness of the proposed compensation mechanism for task-level routing under unified inference.

\section{Shared Semantic Reference Analysis}\label{Semantic}

\subsection{Anchor preservation across tasks.}

\begin{table}[h]
\centering
\caption{Anchor preservation across tasks measured by the average cosine similarity between the final text features and the originally cached text anchors.}
\label{tab:anchor_preservation}
\begin{tabular}{lc}
\toprule
Dataset & Average Cosine Similarity \\
\midrule
CIFAR100   & 0.9987 \\
ImageNet-R & 0.9933 \\
ImageNet100 & 0.9969 \\
\bottomrule
\end{tabular}
\end{table}

To examine whether the shared text branch maintains a stable semantic reference throughout incremental learning, we further compute the average cosine similarity between the final text features of historical classes and their originally cached text anchors. Specifically, for each historical class \(c\), let \(\mathbf{z}_{c}\) denote its cached text anchor at the time when it is first learned, and let \(\mathbf{t}_{c}^{T}\) denote its text feature re-encoded by the shared text branch after the whole incremental process. We measure the semantic preservation of each class by:
\[
\cos(\mathbf{t}_{c}^{T},\mathbf{z}_{c}),
\]
and report the average value over all classes.

The results are summarized in Table~\ref{tab:anchor_preservation}. The average cosine similarities are 0.9987, 0.9933, and 0.9969 on CIFAR100, ImageNet-R, and ImageNet100, respectively. Such consistently high similarity indicates that the shared text branch preserves the semantic anchors of historical classes remarkably well throughout the incremental process, thereby providing a relatively stable shared semantic reference across tasks. By storing visual knowledge in task-specific modules while maintaining relatively stable textual knowledge in the shared text branch, GR4CIL structurally reduces inter-task interference and thereby alleviates catastrophic forgetting during incremental learning.

\subsection{Text-subspace and  Compensation-space similarity}

\begin{table}[h]
\centering
\caption{Average adjacent-task subspace distances under different separation settings. Here, \emph{w/o} denotes the variant without the anchor and separation losses. $\tau$ is the separation threshold. Lower values indicate more similar subspaces.}
\label{tab:adjacent_subspace_distance}
\begin{tabular}{lccccc}
\toprule
 & \emph{w/o} & $\tau=0.5$ & $\tau=0.6$ & $\tau=0.7$ & $\tau=0.8$ \\
\midrule
Compensation-space distance         & 0.8788 & 0.8696 & 0.8624 & 0.8565 & 0.8541 \\
Text-space distance & 0.8860 & 0.9241 & 0.9089 & 0.8895 & 0.8782 \\
\bottomrule
\end{tabular}
\end{table}

To further understand whether the shared text branch provides a relatively consistent semantic reference across tasks, we analyze the similarity between adjacent tasks in both the text subspace and the compensation subspace. Specifically, for each task, we extract orthonormal bases from the task-specific text features and compensation heads via SVD. For two adjacent tasks, we then compute their subspace distance using a symmetric directional distance, defined as the average of the two directional distances in both directions. Concretely, for two bases $B_1$ and $B_2$, the directional distance $d(B_1,B_2)$ measures the average residual norm when each basis vector in $B_1$ is projected onto the subspace spanned by $B_2$. We calculate both $d(B_1,B_2)$ and $d(B_2,B_1)$ and take the average. A smaller value therefore indicates that the two subspaces are more similar. In Table~\ref{tab:adjacent_subspace_distance}, we report the average adjacent-task distances of the text subspace and the compensation subspace under different separation settings.

We compare the variant without the anchor and separation losses (denoted as \emph{w/o}) and the variants using different separation thresholds. The adjacent-task text-space distance gradually decreases as the separation threshold increases, indicating that the shared text branch becomes more consistent across neighboring tasks. This trend is intuitive, since a looser separation constraint makes the learned text subspaces less isolated from each other and therefore more similar across tasks.

A similar tendency can also be observed in the compensation space: as the text-space distance becomes smaller, the adjacent-task compensation-space distance also tends to decrease. This suggests that when different tasks are anchored to a more consistent semantic reference, their compensation heads are more likely to be learned in comparable residual spaces, thereby improving cross-task comparability during unified inference.

At the same time, the \emph{w/o} setting reveals that merely obtaining relatively close text subspaces is not sufficient. Although its text-space distance is already relatively small, the corresponding compensation-space distance does not exhibit the same stable trend as the regularized variants. This suggests that the anchor and separation losses do more than simply reduce semantic drift; they also help regularize inter-class relationships across tasks, which makes the learned compensation spaces more structured and comparable.

Finally, these results also indicate an inherent trade-off. Stronger cross-task semantic consistency usually leads to more similar text and compensation spaces, but overly weak separation may also harm intra-task discrimination and, in turn, affect competition among task branches. Therefore, the separation design should be understood as balancing two objectives: maintaining a stable shared semantic reference across tasks and preserving sufficient task-internal discriminability.

\section{Parameter Analysis}\label{Parameter}

We first analyze the effect of the separation threshold $\tau$, as summarized in Table~\ref{tab:tau_analysis}. As $\tau$ increases from 0.5 to 0.7, Avg-Acc, Last-Acc, and Last-Routing Accuracy all improve consistently; when $\tau$ is further increased to 0.8, the performance drops again. This trend suggests that $\tau$ controls a trade-off between task-internal discriminability and cross-task semantic consistency. A smaller $\tau$ imposes a stronger separation constraint, which improves intra-task discrimination but weakens the semantic consistency of the shared text space across tasks. As a result, the learned compensation spaces become less comparable, which may eventually hurt cross-task competition and routing. In contrast, a larger $\tau$ makes the shared text space more consistent across tasks and thus benefits the comparability of the compensation space, but an overly weak separation constraint may lead to insufficient intra-task discrimination. Therefore, intermediate values such as 0.6 or 0.7 provide a better balance between these two factors, yielding better and more stable classification and routing performance. In our experiments, we use $\tau=0.7$ as the default setting, since it achieves the best Avg-Acc and Last-Acc while matching the best routing accuracy.

\begin{table}[h]
\centering
\caption{Effect of the separation threshold $\tau$ on classification and routing performance on CIFAR-100.}
\label{tab:tau_analysis}
\begin{tabular}{lccc}
\toprule
$\tau$ & Avg-Acc & Last-Acc & Last-Routing Accuracy \\
\midrule
0.5 & 89.06 & 82.44 & 82.82 \\
0.6 & 89.70 & 83.20 & 83.68 \\
0.7 & \textbf{89.77} & \textbf{83.24} & \textbf{83.68} \\
0.8 & 89.24 & 82.80 & 83.18 \\
\bottomrule
\end{tabular}
\end{table}

We further study the effect of the compensation coefficient $\beta$ and the prototype coefficient $\gamma$, as summarized in Table~\ref{tab:beta_gamma_analysis}. When both coefficients are small, the performance is consistently weaker, indicating that neither the compensation term nor the prototype-based confidence cue can be fully utilized. Increasing either $\beta$ or $\gamma$ from 0.1 to 0.2 already leads to clear improvements, showing that both components contribute positively to unified inference.

The best overall performance is achieved at $(\beta,\gamma)=(0.2,0.2)$, suggesting that a moderate and balanced weighting between residual compensation and prototype-based task awareness is most effective. When either weight is further increased to 0.5, the performance drops again. This indicates that overly strong compensation may disturb the relatively stable semantic basis provided by the text classifier, while an excessively large prototype term may overemphasize task-specific distribution cues in the final decision.

\begin{table}[h]
\centering
\caption{Effect of the compensation coefficient $\beta$ and the prototype coefficient $\gamma$ on CIFAR100.}
\label{tab:beta_gamma_analysis}
\begin{tabular}{lccc}
\toprule
$(\beta,\gamma)$ & Avg-Acc & Last-Acc & Last-Routing Accuracy \\
\midrule
$(0.1,0.1)$ & 89.55 & 82.56 & 83.04 \\
$(0.1,0.2)$ & 89.75 & 82.98 & 83.44 \\
$(0.2,0.1)$ & 89.73 & 82.96 & 83.44 \\
$(0.2,0.2)$ & \textbf{89.77} & \textbf{83.24} & \textbf{83.68} \\
$(0.2,0.5)$ & 89.67 & 83.20 & 83.60 \\
$(0.5,0.2)$ & 89.71 & 83.10 & 83.40 \\
$(0.5,0.5)$ & 89.59 & 82.94 & 83.36 \\
\bottomrule
\end{tabular}
\end{table}

\section{Limitation}\label{Limitation}

Although GR4CIL achieves strong performance under the standard CIL setting and naturally leaves a practical interface for extension toward more open scenarios, the current framework still has several limitations. First, the extension interface mainly relies on task-level confidence estimation to determine whether the currently learned knowledge is sufficient to explain an input, and its behavior may therefore still be affected by threshold selection. We thus view it as an extensible entry point beyond standard CIL, rather than a complete solution to open-world inference. Second, the current formulation assumes relatively clear task boundaries during the incremental process. In scenarios where task boundaries are ambiguous, the existing task-specific organization and routing mechanism still require further extension. Nevertheless, its potential OOD-awareness remains closely related to the problem of discovering new tasks under task-agnostic settings, which also suggests a possible direction for future development. Finally, although the proposed method already demonstrates favorable efficiency in terms of parameter count and GPU memory usage, the task-specific visual modules and compensation heads still accumulate as the number of tasks grows. Therefore, further reducing the long-term task-specific overhead remains an important direction for future work.



\end{document}